\documentclass{article}

\usepackage{graphicx} % Required for inserting images

\usepackage{wrapfig}

\usepackage[preprint]{corl_2026} % Uncomment for pre-prints (e.g., arxiv); This is like ``final'', but will remove the CORL footnote.
\usepackage{booktabs}

\usepackage{amsmath}

\usepackage{xcolor} % for the teal color

\title{Cross-Embodiment Robot Manipulation via a Unified Hand Action Space}

\author{
Luis Felipe Casas$^{1}$ \quad
Robert Teal$^{1}$ \quad
Keval Shah$^{1}$ \quad
Abhijit Tadepalli$^{2}$ \\
\textbf{Wanxin Jin}$^{2}$ \quad
\textbf{Yu Xiang}$^{1}$ \\
\\
$^{1}$Intelligent Robotics and Vision Lab, University of Texas at Dallas \\
$^{2}$Intelligent Robotics and Interactive Systems Lab, Arizona State University \\ \\
\texttt{\{Luis.CasasMurillo, Robert.Teal, Keval.Shah, Yu.Xiang\}@utdallas.edu} \\
\texttt{\{vtadepa1, wjin\}@asu.edu} \\
}

\begin{document}
\maketitle

%===============================================================================
\begin{abstract}
Robot manipulation policies are typically tied to specific robotic hand embodiments, limiting the transfer of learned behaviors across platforms with different kinematic structures. In this work, we propose the Unified Hand Action Space (UHAS), a sphere-based unified action representation for cross-embodiment dexterous manipulation. UHAS represents robotic hand actions as geometric deformations of a canonical sphere and uses a Cascade Inverse Kinematics (CIK) algorithm to map the shared representation to embodiment-specific joint configurations. Using reinforcement learning, we train dexterous manipulation policies directly in the proposed action space for in-hand cube reorientation tasks. We evaluate our method in both simulation and real-world experiments across multiple robotic hands, including the Allegro Hand, LEAP Hand, Shadow Hand, and MANO Human Hand. Experimental results demonstrate effective dexterous manipulation, zero-shot transfer to unseen hands, rapid finetuning across embodiments, and successful real-world deployment. Our experiments show that the proposed UHAS representation enables stable dexterous control and cross-embodiment policy transfer across robotic hands.\footnote{Data, code, and videos for the project are available at \url{https://irvlutd.github.io/UHAS/}.}

\end{abstract}

% \begin{abstract}
%     The purpose of this document is to provide both the basic paper template and submission guidelines. Abstracts should be a single paragraph, between 4--6 sentences long, ideally. Gross violations will trigger corrections at the camera-ready phase.
% \end{abstract}

% % Two or three meaningful keywords should be added here
% \keywords{CoRL, Robots, Learning} 

%===============================================================================

\vspace{-2mm}
\section{Introduction}
\vspace{-2mm}
\label{sec:intro}

Robot manipulation has made rapid progress in recent years due to advances in imitation learning~\cite{zhao2023learning,chi2025diffusion}, reinforcement learning~\cite{kalashnikov2018scalable,singh2024dextrah}, and vision-language-action (VLA) models~\cite{kim2024openvla,black2024pi0visionlanguageactionflowmodel,intelligence2025pi05}. While recent policies have demonstrated impressive capabilities in grasping, pick-and-place, dexterous manipulation, and long-horizon task execution, most large-scale robot learning systems and datasets remain dominated by relatively simple end-effectors such as parallel-jaw grippers~\cite{brohan2022rt1,brohan2023rt2visionlanguageactionmodelstransfer,team2024octo,o2024open,khazatsky2024droid}. In contrast, dexterous robotic hands remain comparatively underexplored due to their highly diverse kinematic structures, action parameterizations, and control interfaces. As a result, existing dexterous manipulation systems often rely on embodiment-specific action spaces, datasets, and retraining procedures~\cite{wen2025grdextertechnicalreport,zhang2026dexora,zheng2026egoscale}, making cross-embodiment transfer across robotic hands challenging.

To address this challenge, we investigate the problem of robot learning in a unified hand action space for robot manipulation. \emph{The central idea of our approach is to represent hand actions through the deformation of a canonical sphere representation.} Instead of defining actions directly in embodiment-specific joint spaces, we model manipulation actions as geometric deformations on a shared spherical surface. This representation provides a continuous and embodiment-agnostic interface that captures the semantic behavior of robotic hands while abstracting away low-level hardware differences.

Our sphere-based action representation is motivated by the observation that many robotic hands interact with objects through spatial contact patterns around an object-centered workspace despite their diverse mechanical designs. By representing hand actions as deformations of a canonical sphere, we obtain a shared geometric action space that generalizes across different hand morphologies. Embodiment-specific hand motions are recovered through the proposed inverse kinematics algorithm that maps sphere deformations to executable joint configurations. This unified representation enables cross-embodiment policy transfer, data sharing across heterogeneous robotic platforms, and scalable policy learning for future robot foundation models.

\begin{figure}
 \centering 
 \includegraphics[width=\textwidth ]{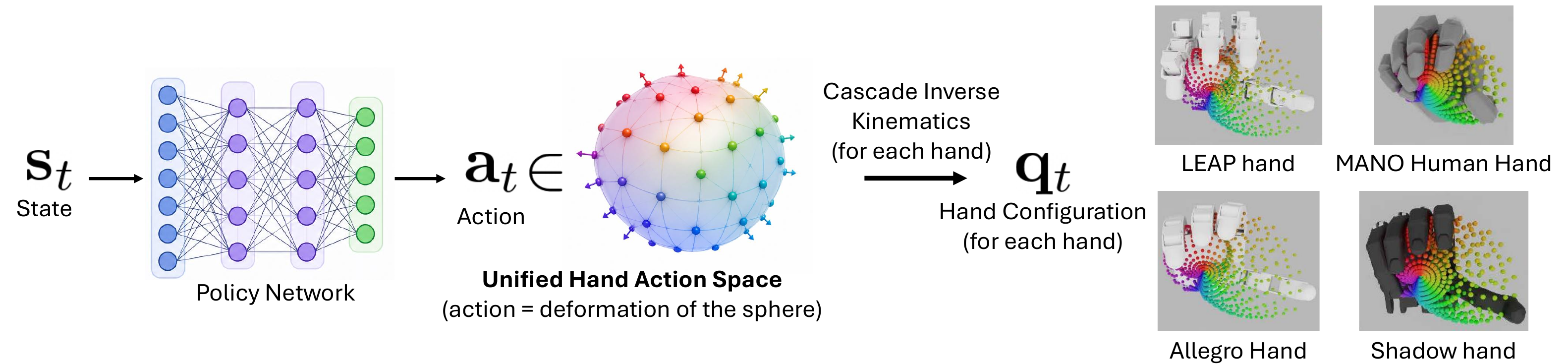}\vspace{-3mm}
   \caption{In our unified hand action space, an action is represented as the deformation of a canonical sphere. A deformed sphere is mapped to hand configurations of various embodiments (LEAP~\cite{shaw2023leaphand}, Allegro~\cite{allegrohand}, MANO Human~\cite{MANO:SIGGRAPHASIA:2017} and Shadow~\cite{shadowhand}).}      
   \label{fig:intro}
   \vspace{-4mm}
\end{figure}

In this work, we build upon the proposed sphere-based unified hand action space to develop a framework for cross-embodiment dexterous manipulation. We introduce a geometric action representation based on sphere deformations together with a cascade inverse kinematics algorithm that maps the unified representation to embodiment-specific hand joint configurations. Using reinforcement learning~\cite{schwarke2025rslrl}, we train manipulation policies directly in the proposed action space for dexterous in-hand manipulation. We evaluate our method on in-hand cube reorientation tasks across multiple robotic hands with different kinematic structures. Experimental results demonstrate that the proposed representation enables stable dexterous control, effective multi-hand policy learning, and meaningful zero-shot transfer to previously unseen robotic hands. The main contributions of this work are:
\begin{itemize}
    \vspace{-2mm}
    \item We propose the Unified Hand Action Space (UHAS), a sphere-based geometric action representation that models robotic hand actions as deformations of a canonical sphere.
    
    \item We develop a Cascade Inverse Kinematics (CIK) algorithm that maps the unified sphere representation to heterogeneous robotic hands with different kinematic structures.
    
    \item We demonstrate dexterous in-hand manipulation, multi-hand policy learning, and cross-embodiment transfer across multiple robotic hands in simulation and the real world, including zero-shot transfer to unseen hands.
\end{itemize}

% \section{Introduction}
	
%     Submission to CoRL 2026 will be entirely electronic, via a web site (not email). Information about the submission process and \LaTeX{} templates are available on the conference web site at \url{https://corl.org/}. For camera ready submission, use the \texttt{final} option for the \texttt{\textbackslash usepackage} command. 

%===============================================================================

% \section{Citations}
% \label{sec:citations}

% 	Citations can be made using either \textbackslash citep\{\} or \textbackslash citet\{\}, depending from the appropriateness. To avoid the citation moving to the next line, it is often a good practice to replace the space before with a tilde (\~{}) character.
% 	Example 1: ``CoRL is the best conference ever~\citep{Gauss1857}.''
% 	Example 2: ``\citet{Lagrange1788} proved, both theoretically and numerically, that CoRL is the best conference ever.''
	
%===============================================================================

\vspace{-4mm}
\section{Related Work}
\vspace{-2mm}
\label{sec:rw}

% \begin{itemize}
%     \item Description of SotA landscape for the solution of the key problem
%     \item Use baselines as main source of references; Reproduce the search tree of related work and search for newer methods for the ones referenced in the baseline (Expand leaves)
%     \item Restate the key difference of our work and a spoiler of good results.
% \end{itemize}

% Go into detail of the current state of the art methods that perform cross embodiment learning, find the most recent and best results and add them here with a brief description. Joint paragraphs for the ones with the same first principles. 

% Expand on intermediate grasp representation

\label{sec:related_work}

\paragraph{Dexterous Manipulation.}
Dexterous in-hand manipulation remains a fundamental challenge in robotics due to the high-dimensional action spaces and complex contact dynamics involved in controlling multi-finger robotic hands. Prior works have demonstrated that reinforcement learning can learn sophisticated dexterous manipulation skills through large-scale simulation training and domain randomization~\cite{andrychowicz2020learning,chen2022towards,handa2023dextreme,ma2024eureka,zakka2025mujocoplayground}. More recent efforts have explored learning dexterous behaviors from demonstrations and large-scale video datasets~\cite{shaw2023videodex,cheng2024open,wang2024dexcap,xu2025dexumi,li2025maniptrans,tao2025dexwild}. In this work, we focus on dexterous in-hand cube reorientation and investigate how a unified geometric action representation can improve manipulation transfer across robotic hands with different kinematic structures.

\vspace{-3mm}
\paragraph{Unified and Cross-Embodiment Action Representations.}
Recent robot learning systems increasingly aim to support multiple robotic embodiments through shared representations and generalist policies. RT-2~\cite{brohan2023rt2visionlanguageactionmodelstransfer}, Octo~\cite{team2024octo}, and OpenVLA~\cite{kim2024openvla} demonstrate the potential of large-scale robot foundation models, but most existing approaches still rely on embodiment-specific action spaces. Several recent works have explored unified representations for cross-embodiment learning, including CrossFormer~\cite{doshi2024scaling}, Universal Actions~\cite{zheng2025universal}, XL-VLA~\cite{jiang2026cross}, and One Hand to Rule Them All~\cite{wei2026one}. These methods use different action heads, latent actions or canonical joint-space representations. Our work introduces the Universal Hand Action Space (UHAS), where manipulation actions are represented as geometric deformations of a canonical sphere shared across robotic hands.

\vspace{-2mm}
\paragraph{Geometric Representations and Cross-Hand Transfer.}
Geometric representations have recently emerged as an effective abstraction for cross-embodiment robotic grasping and manipulation~\cite{wei2024mathcal,khargonkar2024robotfingerprint,fei2025t,wu2025cedex,wu2026dexgrasp,he2026generate}. For example, D(R,O) Grasp~\cite{wei2024mathcal} proposes a unified representation of robot-object interactions for cross-embodiment dexterous grasping, while RobotFingerPrint (RFP)~\cite{khargonkar2024robotfingerprint} establishes dense correspondences between gripper surface points and a canonical sphere for grasp synthesis. Inspired by these geometric representations, we extend the sphere correspondence idea beyond grasp representation and propose the Universal Hand Action Space (UHAS), where sphere deformations themselves define the manipulation action space. 

% To map sphere deformations to executable robot actions, we introduce a Cascade Inverse Kinematics (CIK) algorithm that converts the unified geometric representation into embodiment-specific joint configurations. The proposed approach naturally supports robotic hands with different finger layouts and kinematic structures while enabling cross-embodiment policy transfer.

\vspace{-2mm}
\section{Unified Hand Action Space}
\vspace{-2mm}
\label{sec:method}
% Our method is inspired by the RobotFingerPrint (RFP) intermediate grasp representation~\cite{khargonkar2024robotfingerprint} for robotic grasp synthesis. In RFP, the interior palm surfaces of multiple robotic grippers are mapped using a ``maximal sphere'' representation. Specifically, each point on the gripper's palm interior is assigned spherical coordinates $  (\lambda, \phi)  $, thereby establishing a direct correspondence to a point on the surface of the unit sphere $  S  $, i.e., every palm surface point $  p_{\text{surface}}  $ satisfies $  p_{\text{surface}} \in S  $.

% This mapping is applied to a grasp dataset to train a Conditional Variational Autoencoder (CVAE) generative neural network. The CVAE learns to predict unified gripper coordinates for the surface points of objects, producing maps in which each predicted point corresponds directly to a point on the gripper palm via the shared sphere representation. These predicted maps are subsequently employed in an optimization algorithm to derive a joint configuration $  \mathbf{q}_G  $ of each gripper.

The Unified Hand Action Space (UHAS) is motivated by the central question of whether robotic hand control can be expressed through geometric deformations of a canonical sphere surface. A sphere provides a natural unified representation because it defines a continuous, closed, and topology-agnostic surface that can be consistently parameterized across robotic hands with different kinematic structures and numbers of fingers. Moreover, many grasping and dexterous manipulation behaviors can be interpreted as coordinated radial and angular interactions around an object-centered workspace, making the sphere a convenient geometric abstraction for hand control.

\begin{figure}[h]
    \centering
    \begin{center} 
        \includegraphics[width=\linewidth]{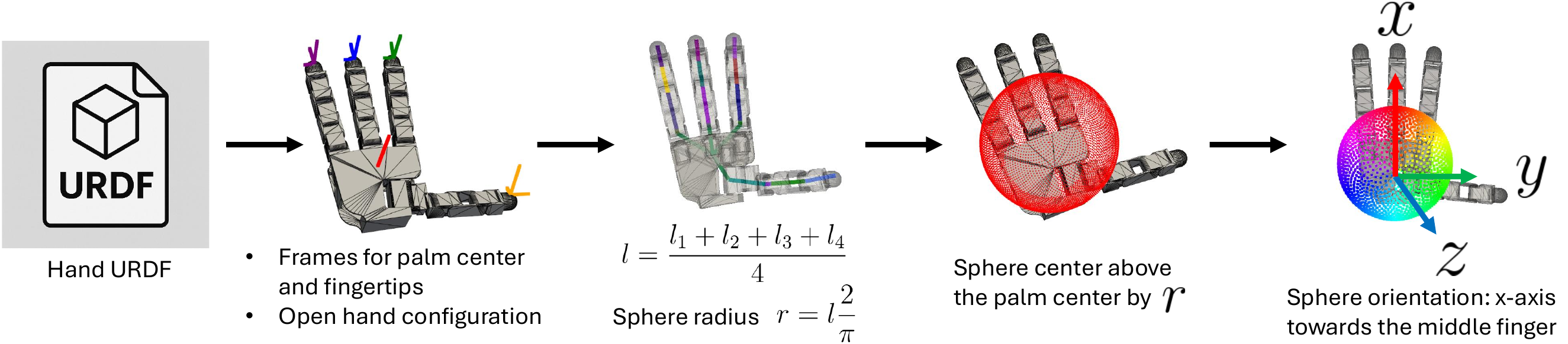}
        \caption{Illustration of the process of creating a sphere for a robotic hand given its URDF.}
    \label{fig:sphere_construction}
    \end{center}
    \vspace{-6mm}
\end{figure}

\subsection{Automatic Sphere Creation for a Robotic Hand}
\label{sec:construction}

Given a robotic hand URDF model, we automatically construct the canonical sphere through kinematic analysis and geometric normalization, as illustrated in Fig.~\ref{fig:sphere_construction}. We first identify the palm and fingertip coordinate frames in an open-hand configuration, where the fingers are fully extended and the fingertip normals approximately align with the palm normal. We then compute the palm center by averaging the finger root positions and measure the average distance $l$ from the palm center to the fingertips. The sphere radius is defined as $r=\frac{2l}{\pi}$, placing the sphere within the natural grasping workspace of the hand and approximately covering a $90^\circ$ arc from the palm center to the fingertips.

Next, the sphere center is positioned along the outward palm normal at a distance $r$ from the palm center, placing the sphere within the natural grasping workspace of the hand where the fingers tend to converge during grasping and dexterous manipulation. To obtain a canonical orientation shared across embodiments, the sphere coordinate frame is defined such that its positive $z$-axis aligns with the outward palm normal and its positive $x$-axis aligns with the middle finger direction. The $y$-axis is then uniquely determined by the resulting $xz$-plane according to the right-hand rule.

To remove embodiment-specific scale differences, all distances in the sphere coordinate frame are normalized by the hand-specific radius $r$. This normalization maps the canonical sphere to a \emph{unit sphere}, producing a scale-invariant representation shared across robotic hands with different kinematic structures and numbers of fingers. The resulting normalized sphere coordinate system defines the Unified Hand Action Space (UHAS), enabling a unified geometric action representation for cross-embodiment policy learning and transfer. Although the UHAS operates on a normalized unit sphere, the corresponding hand-specific sphere parameters are preserved and later used by the Cascade Inverse Kinematics (CIK) algorithm to recover embodiment-specific hand configurations.

\begin{figure}[h]
    \centering
    \begin{center} 
        \includegraphics[width=\linewidth]{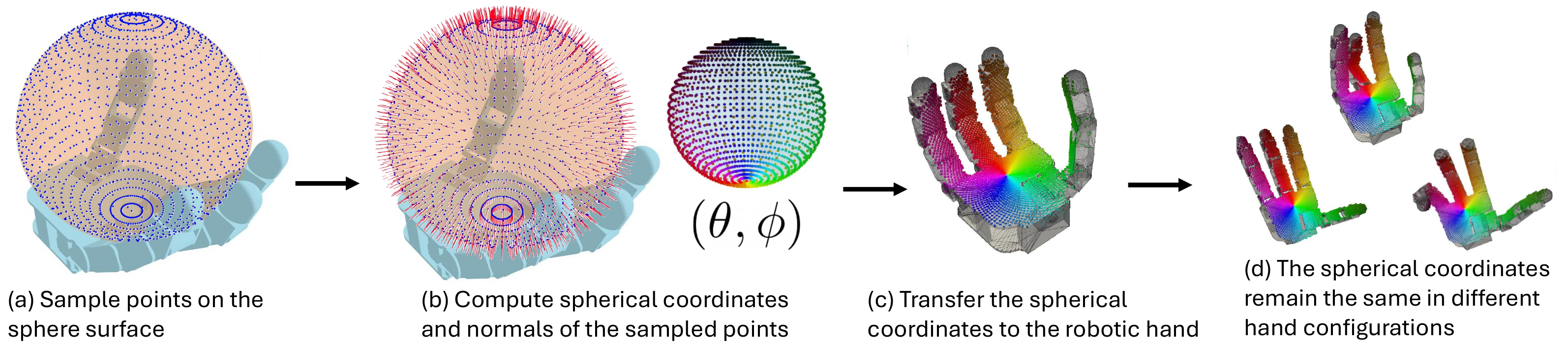}
        \caption{Construction of the unified hand surface correspondence.}
    \label{fig:surface_correspondence}
    \end{center}
    \vspace{-6mm}
\end{figure}

\subsection{Unified Hand Surface Correspondence}
\label{sec:surface_correspondence}

The key idea behind using the sphere for hand control is to establish dense geometric correspondences between the canonical sphere surface and the robotic hand. Once these correspondences are defined, deformations of the sphere can be directly transferred to the hand surface, enabling the sphere to serve as a unified action representation across embodiments.

As illustrated in Fig.~\ref{fig:surface_correspondence}, we first uniformly sample points on the sphere surface, as shown in Fig.~\ref{fig:surface_correspondence}(a). For each sampled point, we compute its spherical coordinates $(\theta,\phi)$ together with its outward surface normal, as illustrated in Fig.~\ref{fig:surface_correspondence}(b), where $\theta$ denotes the azimuthal angle and $\phi$ denotes the polar angle. These spherical coordinates provide a canonical parameterization of the sphere surface independent of any specific robotic hand embodiment. We transfer the spherical coordinates to the robotic hand by projecting sampled sphere points onto nearby points on the interior hand surface, producing dense correspondences between the sphere and the palm and finger surfaces, as illustrated in Fig.~\ref{fig:surface_correspondence}(c). Although the 3D locations of the hand surface points vary under different hand configurations, their associated spherical coordinates remain unchanged, as shown in Fig.~\ref{fig:surface_correspondence}(d). This configuration-invariant parameterization enables different robotic hands to share a common spherical coordinate domain, allowing hand actions to be expressed as geometric deformations of the canonical sphere surface rather than embodiment-specific joint motions.

\subsection{Sphere Deformation Action Space}
\label{sec:sphere_deform}

Directly modeling point-wise deformations of the entire sphere surface is computationally impractical due to the high dimensionality of the resulting action space. Let $(\theta,\phi,r)$ denote the spherical coordinates of a point on the normalized reference sphere, where $\theta$ is the azimuthal angle, $\phi$ is the polar angle, and $r=1$ for the undeformed sphere. To obtain a compact action representation, we model sphere deformations using only $(\Delta\theta,\Delta r)$, corresponding to lateral angular displacements and radial expansions or contractions. This simplification is motivated by the construction of the UHAS, where the north pole of the sphere is rigidly anchored to the palm frame.

%Directly modeling deformations of the entire sphere surface on a point-by-point basis is computationally impractical due to the high dimensionality of the resulting action space. Let $(\theta,\phi,r)$ denote the spherical coordinates of a point on the reference (undeformed) sphere, where $\theta$ is the azimuthal angle, $\phi$ is the polar angle, $r$ is the radial coordinate, and $r=1$ for the undeformed sphere after the radius normalization. In the most general case, a sphere deformation can be represented as perturbations $(\Delta\theta,\Delta\phi,\Delta r)$ applied to every surface point.

%To obtain a compact and tractable action representation, we simplify the deformation model to the pair $(\Delta\theta,\Delta r)$. This simplification is motivated by the construction of the Unified Hand Action Space (UHAS), where the north pole of the sphere is rigidly anchored to the palm frame. As a result, the dominant controllable deformations correspond to lateral angular displacements around the sphere and radial expansions or contractions away from the sphere center.

\begin{figure}[h]
 \centering 
 \includegraphics[width=\textwidth ]{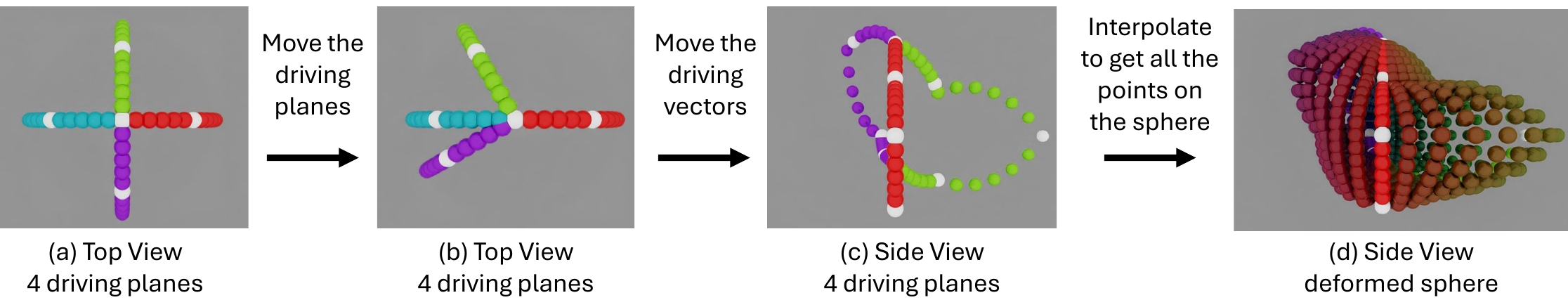}\vspace{-2mm}
   \caption{Sphere deformation parameterization in the Unified Hand Action Space (UHAS). 
(a) Initial configuration of four driving planes. 
(b) Rotating the driving planes controls the lateral deformation $\Delta\theta$. 
(c) Radial displacement of the driving vectors controls $\Delta r$. 
(d) The final deformed sphere reconstructed through interpolation.}
   \label{fig:sphere_deformations}
   \vspace{-2mm}
\end{figure}

We parameterize the sphere deformation using a sparse set of \emph{control primitives}. Specifically, we introduce a finite number of \emph{driving planes}, each defined as a plane passing through the sphere center at a fixed azimuthal angle $\theta_{\text{plane}}$ (Fig.~\ref{fig:sphere_deformations}). These driving planes control the lateral deformation component $\Delta\theta$. Within each driving plane, we further define a discrete set of control points at fixed polar angles $\phi$. The radial displacements of these control points, referred to as \emph{driving vectors}, parameterize the radial deformation component $\Delta r$.

The complete deformation field over the sphere surface is reconstructed through interpolation. First, the $\Delta\theta$ deformation at every surface point is obtained by interpolating the values specified by neighboring driving planes. Next, the $\Delta r$ deformation is computed through a two-dimensional interpolation of the driving-vector displacements in the $(\theta,\phi)$ parameter space. Despite the compact parameterization, this representation can generate a rich set of sphere morphologies suitable for dexterous manipulation. Fig.~\ref{fig:sphere_deformations} illustrates an example of a sphere deformation. \emph{If we use five driving planes with two driving vectors per plane, it results in a 15-dimensional continuous action representation, i.e., $5 + 2 \times 5$.} In practice, we define the driving planes such that they align with the hand fingers.

\subsection{Cascade Inverse Kinematics}
\label{subsec:CIK}

\begin{figure}[h]
    \centering
    \includegraphics[width=\linewidth]{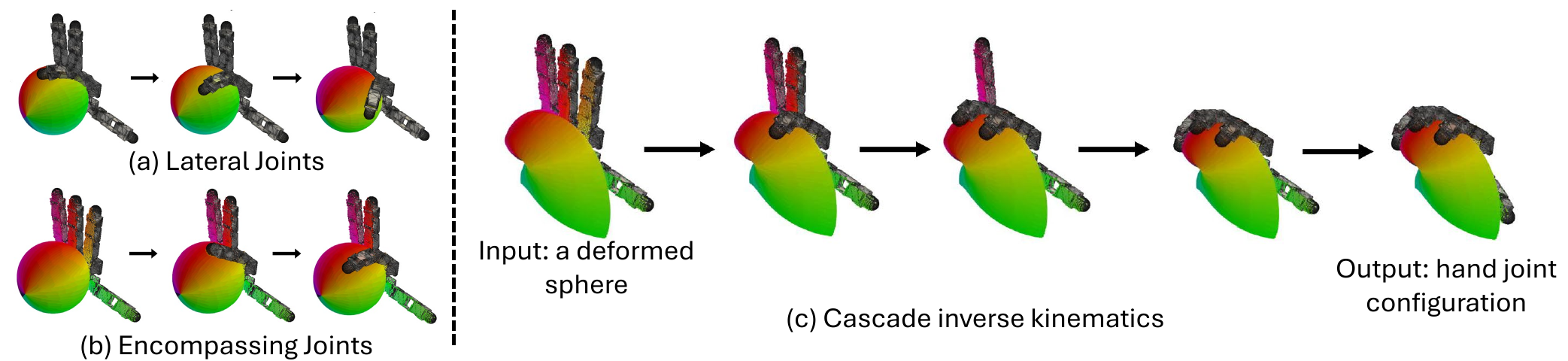}
    \vspace{-2mm}
    \caption{We classify hand joints into (a) lateral joints and (b) encompassing joints and; (c) Illustration of the cascade inverse kinematics algorithm on a deformed sphere.}
    \label{fig:CIK}
    \vspace{-2mm}
\end{figure}

We map deformed sphere surfaces to embodiment-specific hand joint configurations using a novel \emph{Cascade Inverse Kinematics} (CIK) algorithm. Using the surface correspondences defined in Section~\ref{sec:surface_correspondence}, each hand surface point is associated with fixed spherical coordinates $(\theta,\phi)$ on the canonical sphere. Given a deformed sphere, the corresponding target surface positions are obtained directly from the sphere geometry and used by the CIK algorithm to compute the hand joint configuration $\mathbf{q}$.

\vspace{-2mm}
\paragraph{Joint Classification.}
Each hand joint is classified according to its dominant effect on the corresponding hand surface points in the sphere coordinate frame. As illustrated in Fig.~\ref{fig:CIK}(a), lateral joints primarily control the azimuthal angle $\theta$ of the fingertip and finger surface points, producing side-to-side finger motion. In contrast, encompassing joints in Fig.~\ref{fig:CIK}(b) mainly affect the radial distance $r$ and polar angle $\phi$, allowing the fingers to conform to the sphere surface. This decomposition enables the inverse kinematics problem to be solved efficiently in a cascaded manner.

\vspace{-2mm}
\paragraph{Lateral Joint Mapping.}
 
To efficiently control lateral deformations, we precompute a lookup table by sweeping each lateral joint across its full range of motion while re-solving the encompassing joints on the undeformed reference sphere. For every sampled lateral-joint configuration, we record the resulting fingertip azimuthal angle $  \theta  $. The lookup table provides a mapping between lateral joint configurations and $  \theta  $ displacements on the sphere surface. During inference, the target $  \Delta\theta  $ deformation is obtained directly from the displacement of the fingertip's corresponding sphere point computed in Section~\ref{sec:surface_correspondence}. The lateral joint adjustments $  \Delta\mathbf{q}_{\text{lateral}}  $ are then recovered through direct lookup in the precomputed table.

\vspace{-2mm}
\paragraph{Encompassing Cascade.}
After the lateral adjustment stage, the encompassing joints are solved sequentially along the kinematic chain from the finger root to the fingertip. For each encompassing joint, we solve a one-dimensional inverse kinematics subproblem that places the downstream hand surface points onto their corresponding locations on the deformed sphere surface. This cascading procedure progressively conforms the finger geometry to the target sphere deformation while maintaining kinematic feasibility. The resulting joint configuration $\mathbf{q}$ is the final output of the CIK algorithm. The complete procedure is illustrated in Fig.~\ref{fig:CIK}(c). More details of the CIK algorithm are present in Appendix~\ref{appendix:CIK}.

\paragraph{Sphere Controller.}
By combining the sphere deformation parameterization with the proposed Cascade Inverse Kinematics (CIK) algorithm, we obtain a \emph{sphere controller} that maps sphere deformations in the UHAS to embodiment-specific hand joint configurations. During policy execution, the learned policy predicts sphere deformation parameters, which are converted into executable hand motions through CIK. Thanks to its lightweight sequential structure, CIK runs at up to \textbf{150 Hz}, enabling real-time control across diverse robotic hands.

\section{In-hand Manipulation Experiments}
\label{sec:exp}

\begin{figure}
\centering
\includegraphics[width=\linewidth]{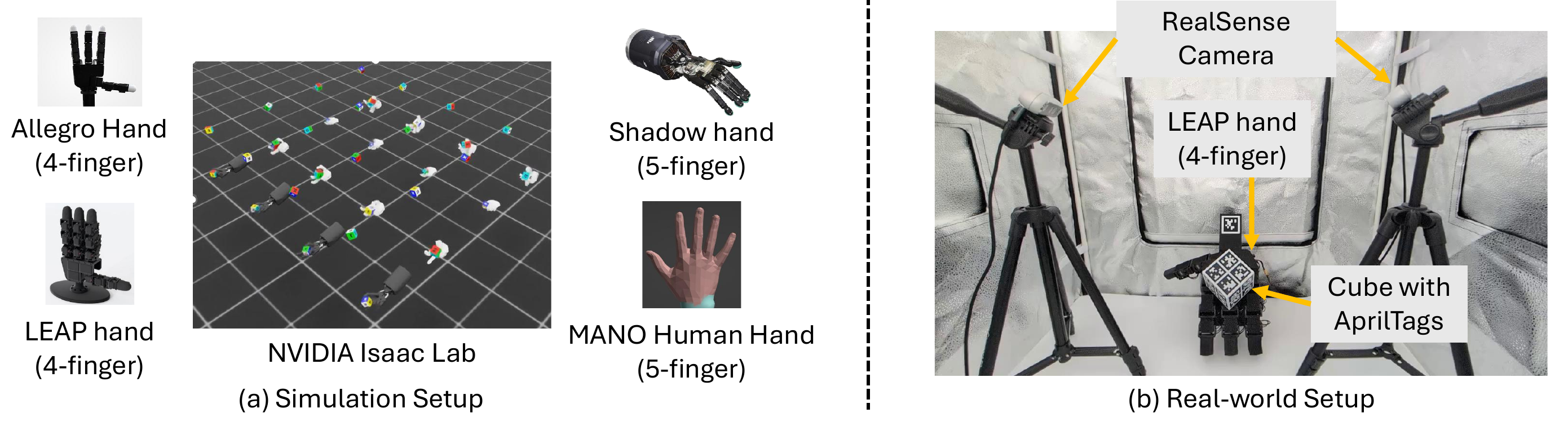}
\vspace{-3mm}
\caption{(a) Simulation setup with 4 hands (b) Our real-world setup of the LEAP hand}
\label{fig:real_world_setup}
\vspace{-4mm}
\end{figure}

\subsection{Experimental Setup}

\paragraph{Tasks.}
We evaluate our method on the task of in-hand cube reorientation in both simulation and the real world. i) For simulation, We use the Repose Cube simulation in NVIDIA Isaac Lab~\cite{mittal2025isaaclab}, and adapt it to use our sphere controller for four hands: Leap Hand~\cite{shaw2023leaphand} and Allegro Hand~\cite{allegrohand} (4 fingers each), and Shadow Hand~\cite{shadowhand} and a simulated MANO human hand model~\cite{MANO:SIGGRAPHASIA:2017} (5 fingers each) as shown in Fig.~\ref{fig:real_world_setup}(a). In each episode in simulation, the robotic hand is required to sequentially reach 10 target cube orientations. If the cube falls, the environment is reset and the remaining reorientation attempts continue until all 10 targets have been evaluated.  ii) In the real world, we conduct experiments with a LEAP hand as shown in Fig.~\ref{fig:real_world_setup}(b). For each trial in the real world, we run the policy until the cube falls. In both simulation and real-world experiments, each target orientation must be reached within 30 seconds. Otherwise, the attempt is considered a failure.

\vspace{-2mm}
\paragraph{Evaluation Metrics.}
We report two primary evaluation metrics. The \textbf{Average Consecutive Reorientations} measures the average number of consecutive successful reorientations achieved before the first cube drop, with a maximum value of 10 in our simulation environment. The \textbf{Success Rate} measures the percentage of successful individual reorientation attempts, where a trial is considered successful if the cube reaches the target orientation without falling. All simulation results are evaluated using 1000 parallel environments. As a baseline, we compare against a policy trained to directly predict hand joint positions while using joint positions and velocities as the input state.

% \begin{figure}[h]
% \centering
% \includegraphics[width=0.25\linewidth]{example-image}
% \caption{Homogeneous observations across Grippers.}
% \label{fig:Homogeneous observations}
% \end{figure}

% \paragraph{Manipulation Policies.}
% We train the in-hand manipulation policies using the RSL-RL implementation of PPO~\cite{schwarke2025rslrl} with a lightweight actor-critic network. The policy outputs actions in the form of sphere deformations, where each finger's lateral movement is controlled by one driving plane with two driving vectors. \yu{Briefly mention how to deal with both 4 and 5 fingers? So we can use the same action for all hands.} These vectors directly influence the encompassing joints through the CIK algorithm. We adopt the same reward formulation as the original Isaac Lab~\cite{mittal2025isaaclab} environment and use identical reward settings for all models and baselines to ensure fair comparisons. For observations, we use homogeneous observations expressed in the canonical sphere coordinate frame and normalized by the corresponding hand-specific sphere radius. This is necessary because raw joint observations are incompatible across robotic hands with different kinematic structures and numbers of fingers. Additional training and observation details are provided in the Appendix~\ref{appendix:training}.

\vspace{-2mm}
\paragraph{Manipulation Policies.}
We train in-hand manipulation policies using the RSL-RL implementation of PPO~\cite{schwarke2025rslrl} with a lightweight actor-critic network. The policy outputs sphere deformations, where each finger is controlled by one driving plane with two driving vectors. To support both 4-finger and 5-finger hands within a unified action space, we add an extra driving plane at the ring-finger position for 4-finger embodiments. We adopt the same reward formulation as the original Isaac Lab~\cite{mittal2025isaaclab} environment and use identical reward settings for all models and baselines. For observations, we use homogeneous observations expressed in the canonical sphere coordinate frame and normalized by the corresponding hand-specific sphere radius, enabling a shared observation space across robotic hands with different kinematic structures and numbers of fingers. Together with the unified action representation, \emph{this allows a single policy to operate across all robotic hands used in our experiments.} Additional training details are provided in Appendix~\ref{appendix:training}.

\subsection{Simulation Results}

\begin{table}[h]
\centering
\caption{
In-hand cube reorientation results. Metrics are reported as (Success Rate / Average Consecutive Reorientations). 
\textbf{Single-Hand}: trained only on the target hand.
\textbf{Joint Control}: baseline joint-space controller trained on the target hand.
\textbf{Multi-Hand}: trained on all four hands.
\textbf{Zero-shot}: trained on all hands except the target hand.
}
\label{tab:main_results}

\setlength{\tabcolsep}{4pt}
\scalebox{0.8}{
\begin{tabular}{lcccc}
\toprule
\textbf{Test Hand} & \textbf{Single-Hand} & \textbf{Joint Control} & \textbf{Multi-Hand} & \textbf{Zero-shot} \\
\midrule
Allegro 
& \textbf{99.1 / 9.6 $\pm$ 1.7} 
& 98.5 / 9.2 $\pm$ 2.2
& 99.2 / 9.5 $\pm$ 1.9
& 95.3 / 7.7 $\pm$ 3.4\\

LEAP 
& \textbf{99.7 / 9.8 $\pm$ 1.1} 
& 98.6 / 9.3 $\pm$ 1.2
& 99.1 / 9.5 $\pm$ 1.9
& 95.5 / 7.7 $\pm$ 3.5 \\

Shadow 
& \textbf{99.3 / 9.6 $\pm$ 1.6} 
& 98.0 / 9.1 $\pm$ 1.9
& 98.7 / 9.2 $\pm$ 2.3
& 85.7 / 4.4 $\pm$ 3.7\\

MANO 
& \textbf{99.8 / 9.9 $\pm$ 1.0} 
& 99.6 / 9.8 $\pm$ 1.4
& 99.5 / 9.8 $\pm$ 1.2
& 98.1 / 8.9 $\pm$ 2.6\\

\bottomrule
\end{tabular}
}
\vspace{-2mm}
\end{table}

\paragraph{Main Results.}
Table~\ref{tab:main_results} evaluates the proposed sphere controller across four robotic hands on the in-hand cube reorientation task. We compare four training settings: \emph{Single-Hand}, where a separate policy is trained for each hand individually; \emph{Joint Control}, a baseline controller operating directly in joint space; \emph{Multi-Hand}, where a single policy is jointly trained across all hands using the proposed UHAS representation; and \emph{Zero-shot}, where the target hand is excluded during training and evaluated without finetuning.

The proposed UHAS representation achieves consistently strong performance across all hands and generally outperforms the joint-space baseline in both task success and long-horizon stability. Importantly, the Multi-Hand results show that a single shared policy can achieve performance comparable to embodiment-specific policies, while the Zero-shot results demonstrate meaningful transfer to previously unseen robotic hands despite substantial differences in kinematic structure and finger morphology. These results highlight the effectiveness of the proposed UHAS representation for cross-embodiment dexterous manipulation.

\vspace{-2mm}
\paragraph{Cross-Morphology Generalization.}
Table~\ref{tab:finger_generalization} evaluates transfer across robotic hands with different numbers of fingers. We train one policy on the two 5-finger hands (MANO and Shadow) and another on the two 4-finger hands (Allegro and LEAP), then evaluate both policies on all four hands. During deployment, models are applied directly without additional processing, where 5-finger models use one driving plane with its vectors per finger, while 4-finger models duplicate the ring-finger plane and the corresponding observations. The results show strong in-distribution performance and meaningful transfer across hand morphologies. Although a performance gap remains compared to in-distribution performance, the transferred policies achieve substantial success on unseen hands, demonstrating that the proposed UHAS representation generalizes across different finger counts and kinematic structures.

\vspace{-2mm}
\paragraph{Finetuning on Unseen Hands.}
Table~\ref{tab:finetuning} evaluates adaptation from the MANO hand to unseen robotic hands. Starting from a policy trained solely on MANO, we finetune the policy on each target hand for only 500 iterations, compared to approximately 4,500 iterations required for training from scratch. Despite this small finetuning budget, the success rate improves substantially across all target hands, recovering strong manipulation performance. These results demonstrate that the proposed UHAS representation enables both meaningful zero-shot transfer and rapid adaptation to new robotic hands.

\begin{table*}[t]
\centering

\begin{minipage}[t]{0.48\textwidth}
\centering
\caption{
Cross-morphology generalization across robotic hands with different numbers of fingers. 
Metrics are reported as (Success Rate / Average Consecutive Reorientations).
}
\label{tab:finger_generalization}

\scriptsize
\setlength{\tabcolsep}{3pt}

\begin{tabular}{lcc}
\toprule
\textbf{Test Hand} 
& \textbf{Train: Shadow + MANO} 
& \textbf{Train: Allegro + LEAP} \\
& \textbf{(5-finger)} 
& \textbf{(4-finger)} \\
\midrule

Allegro (4F) 
& 66.2 / 1.9 $\pm$ 1.9
& \textbf{99.7 / 9.8 $\pm$ 1.2}  \\

LEAP (4F) 
& 80.8 / 3.7 $\pm$ 3.6
& \textbf{99.8 / 9.9 $\pm$ 0.98} \\

Shadow (5F) 
& \textbf{98.6 / 9.3 $\pm$ 2.1} 
& 83.2 / 4.0 $\pm$ 4.0 \\

MANO (5F) 
& \textbf{99.7 / 9.8 $\pm$ 1.3} 
& 95.0 / 7.6 $\pm$ 3.5\\

\bottomrule
\end{tabular}
\end{minipage}
\hfill
\begin{minipage}[t]{0.48\textwidth}
\centering
\caption{
Fast adaptation from MANO to unseen robotic hands. 
The policy is first trained on MANO and finetuned on the target hand for only 500 iterations. 
Metrics are Success Rate / Average Consecutive Reorientations.
}
\label{tab:finetuning}

\scriptsize
\setlength{\tabcolsep}{10pt}

\begin{tabular}{lcc}
\toprule
\textbf{Target Hand} 
& \textbf{Zero-shot} 
& \textbf{+500 Iter} \\
\midrule

Allegro 
& 95.3 / 7.7 $\pm$ 3.4
& \textbf{96.3 / 8.1 $\pm$ 3.3} \\

LEAP 
& 95.5 / 7.7 $\pm$ 3.5
& \textbf{96.2 / 8.0 $\pm$ 3.3} \\

Shadow 
& 85.7 / 4.4 $\pm$ 3.7
& \textbf{95.8 / 7.8 $\pm$ 3.5} \\

\bottomrule
\end{tabular}
\end{minipage}
\vspace{-4mm}
\end{table*}

% \begin{table}[h!]
%     \centering
%     \caption{Generalization across number of fingers (In-Dist vs Out-of-Dist per model). \yu{Write down the training and testing hands in the table. So people can clearly see it.}} 
%     \label{tab:finger_generalization}
%     \begin{tabular}{lccc}
%         \toprule
%         \textbf{Gripper} & \textbf{5-Finger Model} & \textbf{4-Finger Model}\\
%         \midrule
%         Allegro (4-finger) & 61.08 / 1.50  & 99.83 / 9.90\\
%         Leap (4-finger)    & 73.88 / 2.25  & 99.97 / 9.74\\
%         Shadow (5-finger)  & 99.10 / 9.50  & 77.38 / 3.20\\
%         MANO (5-finger)    & 99.65 / 9.78  & 93.20 / 6.25\\
%         \bottomrule
%     \end{tabular}
% \end{table}

% \begin{table}[h!]
%     \centering
%     \caption{Finetuning from MANO (500 iterations) on target grippers.}
%     \label{tab:finetuning}
%     \begin{tabular}{lcc}
%         \toprule
%         \textbf{Target Gripper} & \textbf{Finetuned} & \textbf{MANO}\\
%         \midrule
%         Allegro Hand      & 99.25 / 9.56 & 99.25 / 9.56 \\
%         Leap Hand         & 96.56 / 7.95 & 99.52 / 9.77 \\
%         Shadow Hand       & 97.10 / 7.99 & 99.85 / 9.88 \\
%         \bottomrule
%     \end{tabular}
% \end{table}

\begin{wraptable}{r}{0.6\textwidth}
%\vspace{-5mm}
\centering
\caption{
Ablation on the number of driving vectors per driving plane.
}
\label{tab:vector_ablation}

\scriptsize
\setlength{\tabcolsep}{3pt}

\begin{tabular}{lcccc}
\toprule
\textbf{\# Driving Vectors} 
& \textbf{1} 
& \textbf{2} 
& \textbf{3} 
& \textbf{4} \\
\midrule

\textbf{Success Rate} 
& 98.0 
& 98.7
& \textbf{99.5} 
& 98.1 \\

\textbf{\# Reorientations} 
& 8.8 $\pm$ 2.6
& 9.3 $\pm$ 2.0
& \textbf{9.6 $\pm$ 1.5} 
& 9.1 $\pm$ 2.4 \\

\textbf{Training Time (h)} 
& 5.3 
& \textbf{4.5}
& 6.5
& 5.5 \\

\bottomrule
\end{tabular}

\vspace{-3mm}
\end{wraptable}

% \paragraph{Ablation Studies.}
% We ablate the number of vectors per finger used in the driving planes of the UHAS model. Separate models were trained using 1, 2, 3, and 4 vectors while keeping all other components fixed. All ablation models were trained with lowered randomization values. As shown in Table~\ref{tab:vector_ablation}, using 3 vectors achieves the highest performance, while 2 vectors offers comparable results at a significantly lower training cost. We therefore select the 2-vector configuration for our final model, as the training time advantage becomes even more pronounced when higher levels of randomization are applied. All models were trained using all available hands. Additional ablation studies are provided in Appendix~\ref{appendix:ablations}.  \yu{Add more details about training time}

\vspace{-2mm}
\paragraph{Ablation Studies.}
We ablate the number of driving vectors per finger in the UHAS model by training policies with 1, 2, 3, and 4 vectors while keeping all other components fixed. All four hands are used for training on one NVIDIA A5000 GPU. The training time is measured as the number of iterations required to reach 90\% of the maximum average consecutive reorientations (10). As shown in Table~\ref{tab:vector_ablation}, 2-vectors achieves comparable performance to 3-vectors while requiring the shortest training time. In contrast, 1-vector provides insufficient control flexibility whose training time is longer than 2-vectors , while 4-vectors increase action-space complexity with limited performance gains. We therefore use the 2-vector configuration in our final model. Additional ablation studies are provided in Appendix~\ref{appendix:ablations}.

% \begin{table}[h!]
%     \centering
%     \caption{Ablation on the number of vectors per finger in the driving planes.}
%     \label{tab:vector_ablation}
%     \begin{tabular}{lcccc}
%         \toprule
%         Number of Vectors & \textbf{1} & \textbf{2} & \textbf{3} & \textbf{4} \\
%         \midrule
%         Performance          & 98.03 / 8.93 & 99.30 / 9.64 & 99.46 / 9.70 & 98.83 / 9.31 \\
%         Training Iterations        & 4{,}000      & 3{,}250      & 4{,}250      & 4{,}000      \\
%         \bottomrule
%     \end{tabular}
% \end{table}

\subsection{Real-World Results with Sim-to-Real Transfer}
\label{subsec:real_world}

\begin{table}[h]
    \centering
    \caption{
Real-world in-hand cube reorientation on the LEAP Hand over 10 independent trials. Entries report the number of consecutive successful reorientations before failure.
}
    \label{tab:real_world}
    \scalebox{0.8}{
    \begin{tabular}{lccccccccccc}
        \toprule
        \textbf{Method} & \textbf{1} & \textbf{2} & \textbf{3} & \textbf{4} &
        \textbf{5} &
        \textbf{6} &
        \textbf{7} &
        \textbf{8} &
        \textbf{9} &
        \textbf{10} &
        \textbf{MEAN} \\
        \midrule
        Baseline (Joint Control) & 0 & 0 & 1 & 2 & 0 & 0 & 0 & 1 & 2 & 0 & 0.6 \\

        UHAS (Zero-Shot) & 2 & 4 & 2 & 0 & 1 & 0 & 0 & 0 & 0 & 0 & 0.9     \\        

    UHAS (Trained on Multi-Hand) & 3 & 0 & 1 & 0 & 0 & 5 & 0 & 0 & 0 & 2 & 1.1 \\         
        
        UHAS (Trained on LEAP Hand) & 0 & 2 & 1 & 0 & 1 & 6 & 2 & 1 & 2 & 5 & \textbf{2.0} \\

        % UGAS (Trained on All Grippers)  & 1.6 & 61.54 \\ 

        \bottomrule
    \end{tabular}
    }
\end{table}

\begin{figure}[h]
\centering
\includegraphics[width=\linewidth]{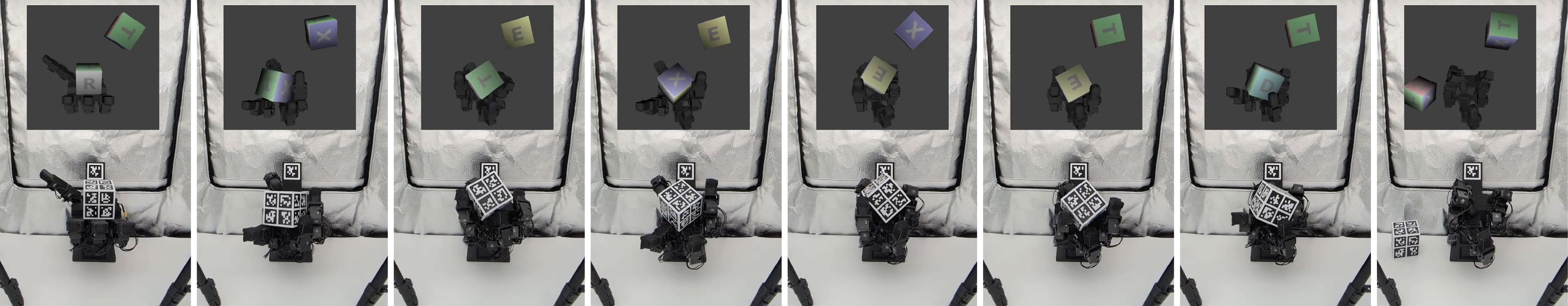}
\caption{An example of a real-world run of our policy for in-hand cube reorientation.}
\label{fig:real_world_example}
\vspace{-2mm}
\end{figure}

We deploy our method on a physical LEAP Hand~\cite{shaw2023leaphand} along with a cube pose estimator based on AprilTags~\cite{apriltag,park2026aprilcube} (Fig.~\ref{fig:real_world_example}). Details of our cube pose estimation algorithm are presented in Appendix~\ref{appendix:cube_pose}. To reduce the sim-to-real gap, we performed system identification on the entire hardware setup with details provided in Appendix~\ref{appendix:sys_ID}. We evaluate in-hand cube reorientation over 10 independent trials per method. In each trial, the policy runs continuously until the cube falls off the hand. Table~\ref{tab:real_world} reports the number of consecutive successful reorientations achieved before failure in each trial, along with the mean across trials. 

We observe a significant performance gap between simulation and the real world despite our efforts on system identification and domain randomization. However, all variants of our method outperform the joint-control baseline. Notably, even the zero-shot UHAS model achieved higher average consecutive reorientations than the baseline. We further observe that the model trained exclusively on the LEAP Hand achieves the best real-world performance (mean of 2.0). Interestingly, the model trained on multiple hands underperforms the single-hand LEAP model. We attribute this to the fact that single-hand training allows the policy to exploit the specific workspace and kinematics of the LEAP Hand, whereas multi-hand training forces the network to learn only movements that are feasible and safe across all hands. Given the relatively simple actor-critic architecture, this results in a more conservative policy when trained on multiple embodiments. Please see the supplementary video for the real-world deployment, and additional results in Appendix~\ref{appendix:results}.

% To facilitate object pose tracking, we utilize custom 3D printed AprilTag markers built into the cube based on a modified aprilcube design. (cite) Previous attempts with printed stickers attached to the cube proved to be too fragile for constant contact with the hand. A RealSense D435 camera mounted on a static tripod tracks these markers alongside a reference AprilTag affixed to the leap hand's base. By computing transformations between the camera, the base, and the object, we map the cube's pose from the camera coordinate frame into the sphere frame for the model observations.

% \begin{table}[h!]
%     \centering
%     \caption{Real-world cube reposing on the LEAP Hand (10 independent trials).}
%     \label{tab:real_world}
%     \begin{tabular}{lcc}
%         \toprule
%         \textbf{Method} & \textbf{Mean Successful Reposes} & \textbf{Success Rate (\%)} \\
%         \midrule
%         UGAS (Trained on LEAP only)     & 5.90 & 83.96 \\
%         UGAS (LEAP OOD)     & 2.0 & 66.66 \\
%         % UGAS (Trained on All Grippers)  & 1.6 & 61.54 \\ 
%         UGAS (Trained on All Grippers)  & 2.8 & 73.68 \\ 
%         Baseline (Isaac Sim Joints)     & 0.70 & 41.18  \\
%         \bottomrule
%     \end{tabular}
% \end{table}

\vspace{-2mm}
\section{Conclusion}
\vspace{-2mm}

In this work, we introduced the Unified Hand Action Space (UHAS), a sphere-based geometric action representation for dexterous manipulation across diverse robotic hands. By establishing dense correspondences between robotic hand surfaces and a canonical sphere representation, the proposed approach enables policies to operate in a shared action space independent of embodiment-specific joint parameterizations. We further proposed the Cascade Inverse Kinematics (CIK) algorithm to map sphere deformations to executable hand motions. Using reinforcement learning, we demonstrated dexterous in-hand cube reorientation across multiple robotic hands in both simulation and the real world, and showed meaningful cross-embodiment transfer through multi-hand, zero-shot, and finetuning experiments.

\vspace{-2mm}
\paragraph{Limitations.}
Dexterous manipulation performance remains highly sensitive to low-level PD controller parameters, requiring extensive domain randomization for robust transfer across robotic hands. In addition, the in-hand cube reorientation task is sensitive to reward design and reinforcement learning hyperparameters. Finally, although UHAS enables transfer across diverse embodiments, performance degrades when transferring between substantially different hand morphologies, such as 4-finger and 5-finger hands.

\acknowledgments{
This work was supported in part by the National Science Foundation (NSF) under Grant Nos. 2346528 and 2520553, the NVIDIA Academic Grant Program Award, and a gift funding from XPeng.}

% \acknowledgments{If a paper is accepted, the final camera-ready version will (and probably should) include acknowledgments. All acknowledgments go at the end of the paper, including thanks to reviewers who gave useful comments, to colleagues who contributed to the ideas, and to funding agencies and corporate sponsors that provided financial support.}

%===============================================================================

% no \bibliographystyle is required, since the corl style is automatically used.
\bibliography{main}  % .bib

@misc{apriltag,
  title        = {AprilTag},
  author       = {{AprilRobotics}},
  howpublished = {\url{https://github.com/AprilRobotics/apriltag}}
}

@software{park2026aprilcube,
  title={AprilCube: 3D-Printable Fiducial Targets for Reliable 6-DoF Pose Estimation},
  author={Park, Younghyo and Agrawal, Pulkit},
  year={2026},
  url={https://github.com/younghyopark/aprilcube},
}

@misc{allegrohand,
  title        = {Allegro Hand},
  author       = {{Wonik Robotics Co., Ltd.}},
  howpublished = {\url{https://www.allegrohand.com/}},
  note         = {Accessed: 2025-08-08}
}

@article{MANO:SIGGRAPHASIA:2017,
      title = {Embodied Hands: Modeling and Capturing Hands and Bodies Together},
      author = {Romero, Javier and Tzionas, Dimitrios and Black, Michael J.},
      journal = {ACM Transactions on Graphics, (Proc. SIGGRAPH Asia)},
      volume = {36},
      number = {6},
      series = {245:1--245:17},
      month = nov,
      year = {2017},
      month_numeric = {11}
  }

@misc{shadowhand,
  title        = {Shadow Dexterous Hand},
  author       = {{The Shadow Robot Company}},
  howpublished = {\url{https://www.shadowrobot.com/products/dexterous-hand/}},
  note         = {Accessed: 2025-08-08}
}

@article{schwarke2025rslrl,
  title   = {RSL-RL: A Learning Library for Robotics Research},
  author  = {Schwarke, Clemens and Mittal, Mayank and Rudin, Nikita and Hoeller, David and Hutter, Marco},
  journal = {arXiv preprint arXiv:2509.10771},
  year    = {2025}
}

@misc{zakka2025mujocoplayground,
      title={MuJoCo Playground},
      author={Kevin Zakka and Baruch Tabanpour and Qiayuan Liao and Mustafa Haiderbhai and Samuel Holt and Jing Yuan Luo and Arthur Allshire and Erik Frey and Koushil Sreenath and Lueder A. Kahrs and Carmelo Sferrazza and Yuval Tassa and Pieter Abbeel},
      year={2025},
      eprint={2502.08844},
      archivePrefix={arXiv},
      primaryClass={cs.RO},
      url={https://arxiv.org/abs/2502.08844},
}

@article{shaw2023leaphand,
      title={LEAP Hand: Low-Cost, Efficient, and Anthropomorphic Hand for Robot Learning},
      author={Shaw, Kenneth and Agarwal, Ananye and Pathak, Deepak},
      journal={Robotics: Science and Systems (RSS)},
      year={2023}}

@inproceedings{khargonkar2024robotfingerprint,
title={RobotFingerPrint: Unified Gripper Coordinate Space for Multi-Gripper Grasp Synthesis},
author={Khargonkar, Ninad and Casas, Luis Felipe and  and Prabhakaran, Balakrishnan and Xiang, Yu},
journal={arXiv preprint arXiv:2409.14519},
year={2024}
}

@article{mittal2025isaaclab,
  title={Isaac Lab: A GPU-Accelerated Simulation Framework for Multi-Modal Robot Learning},
  author={Mayank Mittal and Pascal Roth and James Tigue and Antoine Richard and Octi Zhang and Peter Du and Antonio Serrano-Muñoz and Xinjie Yao and René Zurbrügg and Nikita Rudin and Lukasz Wawrzyniak and Milad Rakhsha and Alain Denzler and Eric Heiden and Ales Borovicka and Ossama Ahmed and Iretiayo Akinola and Abrar Anwar and Mark T. Carlson and Ji Yuan Feng and Animesh Garg and Renato Gasoto and Lionel Gulich and Yijie Guo and M. Gussert and Alex Hansen and Mihir Kulkarni and Chenran Li and Wei Liu and Viktor Makoviychuk and Grzegorz Malczyk and Hammad Mazhar and Masoud Moghani and Adithyavairavan Murali and Michael Noseworthy and Alexander Poddubny and Nathan Ratliff and Welf Rehberg and Clemens Schwarke and Ritvik Singh and James Latham Smith and Bingjie Tang and Ruchik Thaker and Matthew Trepte and Karl Van Wyk and Fangzhou Yu and Alex Millane and Vikram Ramasamy and Remo Steiner and Sangeeta Subramanian and Clemens Volk and CY Chen and Neel Jawale and Ashwin Varghese Kuruttukulam and Michael A. Lin and Ajay Mandlekar and Karsten Patzwaldt and John Welsh and Huihua Zhao and Fatima Anes and Jean-Francois Lafleche and Nicolas Moënne-Loccoz and Soowan Park and Rob Stepinski and Dirk Van Gelder and Chris Amevor and Jan Carius and Jumyung Chang and Anka He Chen and Pablo de Heras Ciechomski and Gilles Daviet and Mohammad Mohajerani and Julia von Muralt and Viktor Reutskyy and Michael Sauter and Simon Schirm and Eric L. Shi and Pierre Terdiman and Kenny Vilella and Tobias Widmer and Gordon Yeoman and Tiffany Chen and Sergey Grizan and Cathy Li and Lotus Li and Connor Smith and Rafael Wiltz and Kostas Alexis and Yan Chang and David Chu and Linxi "Jim" Fan and Farbod Farshidian and Ankur Handa and Spencer Huang and Marco Hutter and Yashraj Narang and Soha Pouya and Shiwei Sheng and Yuke Zhu and Miles Macklin and Adam Moravanszky and Philipp Reist and Yunrong Guo and David Hoeller and Gavriel State},
  journal={arXiv preprint arXiv:2511.04831},
  year={2025},
  url={https://arxiv.org/abs/2511.04831}
}

@string{ICRA = "International Conference on Robotics and Automation"}

@string{CoRL = "Conference on Robot Learning"}

@inproceedings{brohan2022rt1,
      title={RT-1: Robotics Transformer for Real-World Control at Scale}, 
      author={Anthony Brohan and Noah Brown and Justice Carbajal and Yevgen Chebotar and Joseph Dabis and Chelsea Finn and Keerthana Gopalakrishnan and Karol Hausman and Alex Herzog and Jasmine Hsu and Julian Ibarz and Brian Ichter and Alex Irpan and Tomas Jackson and Sally Jesmonth and Nikhil J Joshi and Ryan Julian and Dmitry Kalashnikov and Yuheng Kuang and Isabel Leal and Kuang-Huei Lee and Sergey Levine and Yao Lu and Utsav Malla and Deeksha Manjunath and Igor Mordatch and Ofir Nachum and Carolina Parada and Jodilyn Peralta and Emily Perez and Karl Pertsch and Jornell Quiambao and Kanishka Rao and Michael Ryoo and Grecia Salazar and Pannag Sanketi and Kevin Sayed and Jaspiar Singh and Sumedh Sontakke and Austin Stone and Clayton Tan and Huong Tran and Vincent Vanhoucke and Steve Vega and Quan Vuong and Fei Xia and Ted Xiao and Peng Xu and Sichun Xu and Tianhe Yu and Brianna Zitkovich},
      year={2023},
      booktitle={Robotics: Science and Systems (RSS)},
}

@inproceedings{brohan2023rt2visionlanguageactionmodelstransfer,
      title={RT-2: Vision-Language-Action Models Transfer Web Knowledge to Robotic Control}, 
      author={Anthony Brohan and Noah Brown and Justice Carbajal and Yevgen Chebotar and Xi Chen and Krzysztof Choromanski and Tianli Ding and Danny Driess and Avinava Dubey and Chelsea Finn and Pete Florence and Chuyuan Fu and Montse Gonzalez Arenas and Keerthana Gopalakrishnan and Kehang Han and Karol Hausman and Alexander Herzog and Jasmine Hsu and Brian Ichter and Alex Irpan and Nikhil Joshi and Ryan Julian and Dmitry Kalashnikov and Yuheng Kuang and Isabel Leal and Lisa Lee and Tsang-Wei Edward Lee and Sergey Levine and Yao Lu and Henryk Michalewski and Igor Mordatch and Karl Pertsch and Kanishka Rao and Krista Reymann and Michael Ryoo and Grecia Salazar and Pannag Sanketi and Pierre Sermanet and Jaspiar Singh and Anikait Singh and Radu Soricut and Huong Tran and Vincent Vanhoucke and Quan Vuong and Ayzaan Wahid and Stefan Welker and Paul Wohlhart and Jialin Wu and Fei Xia and Ted Xiao and Peng Xu and Sichun Xu and Tianhe Yu and Brianna Zitkovich},
      year={2023},
      booktitle={Conference on Robot Learning (CoRL)}
}

@article{black2024pi0visionlanguageactionflowmodel,
      title={$\pi_0$: A Vision-Language-Action Flow Model for General Robot Control}, 
      author={Kevin Black and Noah Brown and Danny Driess and Adnan Esmail and Michael Equi and Chelsea Finn and Niccolo Fusai and Lachy Groom and Karol Hausman and Brian Ichter and Szymon Jakubczak and Tim Jones and Liyiming Ke and Sergey Levine and Adrian Li-Bell and Mohith Mothukuri and Suraj Nair and Karl Pertsch and Lucy Xiaoyang Shi and James Tanner and Quan Vuong and Anna Walling and Haohuan Wang and Ury Zhilinsky},
      year={2024},
      journal={arXiv preprint arxiv:2410.24164},
}

@article{intelligence2025pi05,
  title={$\pi_{0.5}$: a Vision-Language-Action Model with Open-World Generalization},
  author = {Kevin Black and Noah Brown and James Darpinian and Karan Dhabalia and Danny Driess and Adnan Esmail and Michael Equi and Chelsea Finn and Niccolo Fusai and Manuel Y. Galliker and Dibya Ghosh and Lachy Groom and Karol Hausman and Brian Ichter and Szymon Jakubczak and Tim Jones and Liyiming Ke and Devin LeBlanc and Sergey Levine and Adrian Li-Bell and Mohith Mothukuri and Suraj Nair and Karl Pertsch and Allen Z. Ren and Lucy Xiaoyang Shi and Laura Smith and Jost Tobias Springenberg and Kyle Stachowicz and James Tanner and Quan Vuong and Homer Walke and Anna Walling and Haohuan Wang and Lili Yu and Ury Zhilinsky},
  journal={arXiv preprint arXiv:2504.16054},
  year={2025}
}

@inproceedings{
kim2024openvla,
title={Open{VLA}: An Open-Source Vision-Language-Action Model},
author={Moo Jin Kim and Karl Pertsch and Siddharth Karamcheti and Ted Xiao and Ashwin Balakrishna and Suraj Nair and Rafael Rafailov and Ethan P Foster and Pannag R Sanketi and Quan Vuong and Thomas Kollar and Benjamin Burchfiel and Russ Tedrake and Dorsa Sadigh and Sergey Levine and Percy Liang and Chelsea Finn},
booktitle={Conference on Robot Learning (CoRL)},
year={2024},
}

@article{zhao2023learning,
  title={Learning fine-grained bimanual manipulation with low-cost hardware},
  author={Zhao, Tony Z and Kumar, Vikash and Levine, Sergey and Finn, Chelsea},
  journal={arXiv preprint arXiv:2304.13705},
  year={2023}
}

@inproceedings{o2024open,
  title={Open {X-E}mbodiment: Robotic Learning Datasets and {RT-X} Models},
author = {Open X-Embodiment Collaboration and Abby O'Neill and Abdul Rehman and Abhinav Gupta and Abhiram Maddukuri and Abhishek Gupta and Abhishek Padalkar and Abraham Lee and Acorn Pooley and Agrim Gupta and Ajay Mandlekar and Ajinkya Jain and Albert Tung and Alex Bewley and Alex Herzog and Alex Irpan and Alexander Khazatsky and Anant Rai and Anchit Gupta and Andrew Wang and Andrey Kolobov and Anikait Singh and Animesh Garg and Aniruddha Kembhavi and Annie Xie and Anthony Brohan and Antonin Raffin and Archit Sharma and Arefeh Yavary and Arhan Jain and Ashwin Balakrishna and Ayzaan Wahid and Ben Burgess-Limerick and Beomjoon Kim and Bernhard Schölkopf and Blake Wulfe and Brian Ichter and Cewu Lu and Charles Xu and Charlotte Le and Chelsea Finn and Chen Wang and Chenfeng Xu and Cheng Chi and Chenguang Huang and Christine Chan and Christopher Agia and Chuer Pan and Chuyuan Fu and Coline Devin and Danfei Xu and Daniel Morton and Danny Driess and Daphne Chen and Deepak Pathak and Dhruv Shah and Dieter Büchler and Dinesh Jayaraman and Dmitry Kalashnikov and Dorsa Sadigh and Edward Johns and Ethan Foster and Fangchen Liu and Federico Ceola and Fei Xia and Feiyu Zhao and Felipe Vieira Frujeri and Freek Stulp and Gaoyue Zhou and Gaurav S. Sukhatme and Gautam Salhotra and Ge Yan and Gilbert Feng and Giulio Schiavi and Glen Berseth and Gregory Kahn and Guangwen Yang and Guanzhi Wang and Hao Su and Hao-Shu Fang and Haochen Shi and Henghui Bao and Heni Ben Amor and Henrik I Christensen and Hiroki Furuta and Homanga Bharadhwaj and Homer Walke and Hongjie Fang and Huy Ha and Igor Mordatch and Ilija Radosavovic and Isabel Leal and Jacky Liang and Jad Abou-Chakra and Jaehyung Kim and Jaimyn Drake and Jan Peters and Jan Schneider and Jasmine Hsu and Jay Vakil and Jeannette Bohg and Jeffrey Bingham and Jeffrey Wu and Jensen Gao and Jiaheng Hu and Jiajun Wu and Jialin Wu and Jiankai Sun and Jianlan Luo and Jiayuan Gu and Jie Tan and Jihoon Oh and Jimmy Wu and Jingpei Lu and Jingyun Yang and Jitendra Malik and João Silvério and Joey Hejna and Jonathan Booher and Jonathan Tompson and Jonathan Yang and Jordi Salvador and Joseph J. Lim and Junhyek Han and Kaiyuan Wang and Kanishka Rao and Karl Pertsch and Karol Hausman and Keegan Go and Keerthana Gopalakrishnan and Ken Goldberg and Kendra Byrne and Kenneth Oslund and Kento Kawaharazuka and Kevin Black and Kevin Lin and Kevin Zhang and Kiana Ehsani and Kiran Lekkala and Kirsty Ellis and Krishan Rana and Krishnan Srinivasan and Kuan Fang and Kunal Pratap Singh and Kuo-Hao Zeng and Kyle Hatch and Kyle Hsu and Laurent Itti and Lawrence Yunliang Chen and Lerrel Pinto and Li Fei-Fei and Liam Tan and Linxi "Jim" Fan and Lionel Ott and Lisa Lee and Luca Weihs and Magnum Chen and Marion Lepert and Marius Memmel and Masayoshi Tomizuka and Masha Itkina and Mateo Guaman Castro and Max Spero and Maximilian Du and Michael Ahn and Michael C. Yip and Mingtong Zhang and Mingyu Ding and Minho Heo and Mohan Kumar Srirama and Mohit Sharma and Moo Jin Kim and Muhammad Zubair Irshad and Naoaki Kanazawa and Nicklas Hansen and Nicolas Heess and Nikhil J Joshi and Niko Suenderhauf and Ning Liu and Norman Di Palo and Nur Muhammad Mahi Shafiullah and Oier Mees and Oliver Kroemer and Osbert Bastani and Pannag R Sanketi and Patrick "Tree" Miller and Patrick Yin and Paul Wohlhart and Peng Xu and Peter David Fagan and Peter Mitrano and Pierre Sermanet and Pieter Abbeel and Priya Sundaresan and Qiuyu Chen and Quan Vuong and Rafael Rafailov and Ran Tian and Ria Doshi and Roberto Mart{'i}n-Mart{'i}n and Rohan Baijal and Rosario Scalise and Rose Hendrix and Roy Lin and Runjia Qian and Ruohan Zhang and Russell Mendonca and Rutav Shah and Ryan Hoque and Ryan Julian and Samuel Bustamante and Sean Kirmani and Sergey Levine and Shan Lin and Sherry Moore and Shikhar Bahl and Shivin Dass and Shubham Sonawani and Shubham Tulsiani and Shuran Song and Sichun Xu and Siddhant Haldar and Siddharth Karamcheti and Simeon Adebola and Simon Guist and Soroush Nasiriany and Stefan Schaal and Stefan Welker and Stephen Tian and Subramanian Ramamoorthy and Sudeep Dasari and Suneel Belkhale and Sungjae Park and Suraj Nair and Suvir Mirchandani and Takayuki Osa and Tanmay Gupta and Tatsuya Harada and Tatsuya Matsushima and Ted Xiao and Thomas Kollar and Tianhe Yu and Tianli Ding and Todor Davchev and Tony Z. Zhao and Travis Armstrong and Trevor Darrell and Trinity Chung and Vidhi Jain and Vikash Kumar and Vincent Vanhoucke and Vitor Guizilini and Wei Zhan and Wenxuan Zhou and Wolfram Burgard and Xi Chen and Xiangyu Chen and Xiaolong Wang and Xinghao Zhu and Xinyang Geng and Xiyuan Liu and Xu Liangwei and Xuanlin Li and Yansong Pang and Yao Lu and Yecheng Jason Ma and Yejin Kim and Yevgen Chebotar and Yifan Zhou and Yifeng Zhu and Yilin Wu and Ying Xu and Yixuan Wang and Yonatan Bisk and Yongqiang Dou and Yoonyoung Cho and Youngwoon Lee and Yuchen Cui and Yue Cao and Yueh-Hua Wu and Yujin Tang and Yuke Zhu and Yunchu Zhang and Yunfan Jiang and Yunshuang Li and Yunzhu Li and Yusuke Iwasawa and Yutaka Matsuo and Zehan Ma and Zhuo Xu and Zichen Jeff Cui and Zichen Zhang and Zipeng Fu and Zipeng Lin},
  booktitle={2024 IEEE International Conference on Robotics and Automation (ICRA)},
  pages={6892--6903},
  year={2024},
  organization={IEEE}
}

@article{khazatsky2024droid,
    title   = {DROID: A Large-Scale In-The-Wild Robot Manipulation Dataset},
    author  = {Alexander Khazatsky and Karl Pertsch and Suraj Nair and Ashwin Balakrishna and Sudeep Dasari and Siddharth Karamcheti and Soroush Nasiriany and Mohan Kumar Srirama and Lawrence Yunliang Chen and Kirsty Ellis and Peter David Fagan and Joey Hejna and Masha Itkina and Marion Lepert and Yecheng Jason Ma and Patrick Tree Miller and Jimmy Wu and Suneel Belkhale and Shivin Dass and Huy Ha and Arhan Jain and Abraham Lee and Youngwoon Lee and Marius Memmel and Sungjae Park and Ilija Radosavovic and Kaiyuan Wang and Albert Zhan and Kevin Black and Cheng Chi and Kyle Beltran Hatch and Shan Lin and Jingpei Lu and Jean Mercat and Abdul Rehman and Pannag R Sanketi and Archit Sharma and Cody Simpson and Quan Vuong and Homer Rich Walke and Blake Wulfe and Ted Xiao and Jonathan Heewon Yang and Arefeh Yavary and Tony Z. Zhao and Christopher Agia and Rohan Baijal and Mateo Guaman Castro and Daphne Chen and Qiuyu Chen and Trinity Chung and Jaimyn Drake and Ethan Paul Foster and Jensen Gao and Vitor Guizilini and David Antonio Herrera and Minho Heo and Kyle Hsu and Jiaheng Hu and Muhammad Zubair Irshad and Donovon Jackson and Charlotte Le and Yunshuang Li and Kevin Lin and Roy Lin and Zehan Ma and Abhiram Maddukuri and Suvir Mirchandani and Daniel Morton and Tony Nguyen and Abigail O'Neill and Rosario Scalise and Derick Seale and Victor Son and Stephen Tian and Emi Tran and Andrew E. Wang and Yilin Wu and Annie Xie and Jingyun Yang and Patrick Yin and Yunchu Zhang and Osbert Bastani and Glen Berseth and Jeannette Bohg and Ken Goldberg and Abhinav Gupta and Abhishek Gupta and Dinesh Jayaraman and Joseph J Lim and Jitendra Malik and Roberto Martín-Martín and Subramanian Ramamoorthy and Dorsa Sadigh and Shuran Song and Jiajun Wu and Michael C. Yip and Yuke Zhu and Thomas Kollar and Sergey Levine and Chelsea Finn},
    year    = {2024},
}

@article{chi2025diffusion,
  title={Diffusion policy: Visuomotor policy learning via action diffusion},
  author={Chi, Cheng and Xu, Zhenjia and Feng, Siyuan and Cousineau, Eric and Du, Yilun and Burchfiel, Benjamin and Tedrake, Russ and Song, Shuran},
  journal={The International Journal of Robotics Research},
  volume={44},
  number={10-11},
  pages={1684--1704},
  year={2025},
  publisher={Sage Publications Sage UK: London, England}
}

@inproceedings{kalashnikov2018scalable,
  title={Scalable deep reinforcement learning for vision-based robotic manipulation},
  author={Kalashnikov, Dmitry and Irpan, Alex and Pastor, Peter and Ibarz, Julian and Herzog, Alexander and Jang, Eric and Quillen, Deirdre and Holly, Ethan and Kalakrishnan, Mrinal and Vanhoucke, Vincent and Levine, Sergey},
  booktitle={Conference on robot learning},
  pages={651--673},
  year={2018},
  organization={PMLR}
}

@article{singh2024dextrah,
  title={Dextrah-rgb: Visuomotor policies to grasp anything with dexterous hands},
  author={Singh, Ritvik and Allshire, Arthur and Handa, Ankur and Ratliff, Nathan and Van Wyk, Karl},
  journal={arXiv preprint arXiv:2412.01791},
  year={2024}
}

@article{andrychowicz2020learning,
  title={Learning dexterous in-hand manipulation},
  author = {Marcin Andrychowicz and Bowen Baker and Maciek Chociej and Rafal J{\'o}zefowicz and Bob McGrew and Jakub Pachocki and Arthur Petron and Matthias Plappert and Glenn Powell and Alex Ray and Jonas Schneider and Szymon Sidor and Josh Tobin and Peter Welinder and Lilian Weng and Wojciech Zaremba},
  journal={The International Journal of Robotics Research},
  volume={39},
  number={1},
  pages={3--20},
  year={2020},
  publisher={SAGE Publications Sage UK: London, England}
}

@inproceedings{ma2024eureka,
  title={Eureka: Human-level reward design via coding large language models},
  author={Ma, Yecheng Jason and Liang, William and Wang, Guanzhi and Huang, De-An and Bastani, Osbert and Jayaraman, Dinesh and Zhu, Yuke and Fan, Jim and others},
  booktitle={International conference on learning Representations},
  volume={2024},
  pages={26516--26560},
  year={2024}
}

@inproceedings{handa2023dextreme,
  title={Dextreme: Transfer of agile in-hand manipulation from simulation to reality},
  author = {Ankur Handa and Arthur Allshire and Viktor Makoviychuk and Aleksei Petrenko and Ritvik Singh and Jingzhou Liu and Denys Makoviichuk and Karl Van Wyk and Alexander Zhurkevich and Balakumar Sundaralingam and Yashraj Narang and Jean-Francois Lafleche and Dieter Fox and Gavriel State},
  booktitle={2023 IEEE International Conference on Robotics and Automation (ICRA)},
  pages={5977--5984},
  year={2023},
  organization={IEEE}
}

@article{chen2022towards,
  title={Towards human-level bimanual dexterous manipulation with reinforcement learning},
  author={Chen, Yuanpei and Wu, Tianhao and Wang, Shengjie and Feng, Xidong and Jiang, Jiechuan and Lu, Zongqing and McAleer, Stephen and Dong, Hao and Zhu, Song-Chun and Yang, Yaodong},
  journal={Advances in Neural Information Processing Systems},
  volume={35},
  pages={5150--5163},
  year={2022}
}

@article{xu2025dexumi,
  title={Dexumi: Using human hand as the universal manipulation interface for dexterous manipulation},
  author={Xu, Mengda and Zhang, Han and Hou, Yifan and Xu, Zhenjia and Fan, Linxi and Veloso, Manuela and Song, Shuran},
  journal={arXiv preprint arXiv:2505.21864},
  year={2025}
}

@inproceedings{shaw2023videodex,
  title={Videodex: Learning dexterity from internet videos},
  author={Shaw, Kenneth and Bahl, Shikhar and Pathak, Deepak},
  booktitle={Conference on Robot Learning},
  pages={654--665},
  year={2023},
  organization={PMLR}
}

@article{cheng2024open,
  title={Open-television: Teleoperation with immersive active visual feedback},
  author={Cheng, Xuxin and Li, Jialong and Yang, Shiqi and Yang, Ge and Wang, Xiaolong},
  journal={arXiv preprint arXiv:2407.01512},
  year={2024}
}

@article{wang2024dexcap,
  title={Dexcap: Scalable and portable mocap data collection system for dexterous manipulation},
  author={Wang, Chen and Shi, Haochen and Wang, Weizhuo and Zhang, Ruohan and Fei-Fei, Li and Liu, C Karen},
  journal={arXiv preprint arXiv:2403.07788},
  year={2024}
}

@inproceedings{li2025maniptrans,
  title={Maniptrans: Efficient dexterous bimanual manipulation transfer via residual learning},
  author={Li, Kailin and Li, Puhao and Liu, Tengyu and Li, Yuyang and Huang, Siyuan},
  booktitle={Proceedings of the IEEE/CVF Conference on Computer Vision and Pattern Recognition},
  pages={6991--7003},
  year={2025}
}

@article{tao2025dexwild,
  title={Dexwild: Dexterous human interactions for in-the-wild robot policies},
  author={Tao, Tony and Srirama, Mohan Kumar and Liu, Jason Jingzhou and Shaw, Kenneth and Pathak, Deepak},
  journal={arXiv preprint arXiv:2505.07813},
  year={2025}
}

@article{team2024octo,
  title={Octo: An open-source generalist robot policy},
  author = {Dibya Ghosh and Homer Walke and Karl Pertsch and Kevin Black and Oier Mees and Sudeep Dasari and Joey Hejna and Tobias Kreiman and Ria Doshi and Charles Xu and Jianlan Luo and You Liang Tan and Lawrence Yunliang Chen and Pannag Sanketi and Quan Vuong and Ted Xiao and Dorsa Sadigh and Chelsea Finn and Sergey Levine},
  journal={arXiv preprint arXiv:2405.12213},
  year={2024}
}

@inproceedings{zheng2025universal,
  title={Universal actions for enhanced embodied foundation models},
  author={Zheng, Jinliang and Li, Jianxiong and Liu, Dongxiu and Zheng, Yinan and Wang, Zhihao and Ou, Zhonghong and Liu, Yu and Liu, Jingjing and Zhang, Ya-Qin and Zhan, Xianyuan},
  booktitle={Proceedings of the Computer Vision and Pattern Recognition Conference},
  pages={22508--22519},
  year={2025}
}

@article{jiang2026cross,
  title={Cross-Hand Latent Representation for Vision-Language-Action Models},
  author={Jiang, Guangqi and Liang, Yutong and Ye, Jianglong and Huang, Jia-Yang and Jing, Changwei and Duan, Rocky and Abbeel, Pieter and Wang, Xiaolong and Zou, Xueyan},
  journal={arXiv preprint arXiv:2603.10158},
  year={2026}
}

@article{wei2026one,
  title={One Hand to Rule Them All: Canonical Representations for Unified Dexterous Manipulation},
  author={Wei, Zhenyu and Yao, Yunchao and Ding, Mingyu},
  journal={arXiv preprint arXiv:2602.16712},
  year={2026}
}

@article{wei2024mathcal,
  title={D (R, O) Grasp: A Unified Representation of Robot and Object Interaction for Cross-Embodiment Dexterous Grasping},
  author={Wei, Zhenyu and Xu, Zhixuan and Guo, Jingxiang and Hou, Yiwen and Gao, Chongkai and Cai, Zhehao and Luo, Jiayu and Shao, Lin},
  journal={arXiv preprint arXiv:2410.01702},
  year={2024}
}

@article{wu2026dexgrasp,
  title={DexGrasp-Zero: A Morphology-Aligned Policy for Zero-Shot Cross-Embodiment Dexterous Grasping},
  author={Wu, Yuliang and Lin, Yanhan and Lao, WengKit and Lin, Yuhao and Wei, Yi-Lin and Zheng, Wei-Shi and Wu, Ancong},
  journal={arXiv preprint arXiv:2603.16806},
  year={2026}
}

@article{wu2025cedex,
  title={CEDex: Cross-Embodiment Dexterous Grasp Generation at Scale from Human-like Contact Representations},
  author={Wu, Zhiyuan and Potamias, Rolandos Alexandros and Zhang, Xuyang and Zhang, Zhongqun and Deng, Jiankang and Luo, Shan},
  journal={arXiv preprint arXiv:2509.24661},
  year={2025}
}

@article{he2026generate,
  title={Generate, Transfer, Adapt: Learning Functional Dexterous Grasping from a Single Human Demonstration},
  author={He, Xingyi and Polavaram, Adhitya and Cao, Yunhao and Deshmukh, Om and Wang, Tianrui and Zhou, Xiaowei and Fang, Kuan},
  journal={arXiv preprint arXiv:2601.05243},
  year={2026}
}

@article{fei2025t,
  title={T (R, O) Grasp: Efficient Graph Diffusion of Robot-Object Spatial Transformation for Cross-Embodiment Dexterous Grasping},
  author={Fei, Xin and Xu, Zhixuan and Fang, Huaicong and Zhang, Tianrui and Shao, Lin},
  journal={arXiv preprint arXiv:2510.12724},
  year={2025}
}

@article{zheng2026egoscale,
  title={Egoscale: Scaling dexterous manipulation with diverse egocentric human data},
  author = {Ruijie Zheng and Dantong Niu and Yuqi Xie and Jing Wang and Mengda Xu and Yunfan Jiang and Fernando Casta{\~n}eda and Fengyuan Hu and You Liang Tan and Letian Fu and Trevor Darrell and Furong Huang and Yuke Zhu and Danfei Xu and Linxi Fan},
  journal={arXiv preprint arXiv:2602.16710},
  year={2026}
}

@article{zhang2026dexora,
  title={Dexora: Open-source VLA for High-DoF Bimanual Dexterity},
  author = {Zongzheng Zhang and Jingrui Pang and Zhuo Yang and Kun Li and Minwen Liao and Saining Zhang and Guoxuan Chi and Jinbang Guo and Huan-ang Gao and Modi Shi and Dongyun Ge and Yao Mu and Jiayuan Gu and Rui Chen and Hao Dong and Huazhe Xu and Li Yi and Yixin Zhu and Hang Zhao and Pengwei Wang and Shanghang Zhang and Guocai Yao and Jianyu Chen and Hongyang Li and Hao Zhao},
  journal={arXiv preprint arXiv:2605.18722},
  year={2026}
}

@article{wen2025grdextertechnicalreport,
    title={GR-Dexter Technical Report},
    author={Wen, Ruoshi and Chen, Guangzeng and Cui, Zhongren and Du, Min and Gou, Yang and Han, Zhigang and Huang, Liqun and Lei, Mingyu and Li, Yunfei and Li, Zhuohang and Liu, Wenlei and Liu, Yuxiao and Ma, Xiao and Niu, Hao and Ouyang, Yutao and Ren, Zeyu and Shi, Haixin and Xu, Wei and Zhang, Haoxiang and Zhang, Jiajun and Zhang, Xiao and Zheng, Liwei and Zhong, Weiheng and Zhou, Yifei and Zhu, Zhengming and Li, Hang},
    journal={arXiv preprint arXiv:2512.24210},
    year={2025}
  }

@article{doshi2024scaling,
  title={Scaling cross-embodied learning: One policy for manipulation, navigation, locomotion and aviation},
  author={Doshi, Ria and Walke, Homer and Mees, Oier and Dasari, Sudeep and Levine, Sergey},
  journal={arXiv preprint arXiv:2408.11812},
  year={2024}
}

\newpage

\appendix

\section{Cascade Inverse Kinematics}
\label{appendix:CIK}

The Cascade Inverse Kinematics (CIK) algorithm maps a deformed canonical sphere to embodiment-specific joint configurations $\mathbf{q}$ for arbitrary robotic hands. Its objective is to position the hand surface points as close as possible to their corresponding locations on the input deformed sphere while respecting kinematic constraints. These correspondences are established during automatic sphere creation and surface projection (Section \ref{sec:surface_correspondence} and Fig. \ref{fig:surface_correspondence}).

Traditional numerical inverse kinematics solvers are computationally expensive and unsuitable for real-time dexterous control. In contrast, CIK exploits the geometric structure of the Unified Hand Action Space together with a cascaded decomposition of joint responsibilities. In our real-world setup, the complete CIK pipeline runs at approximately \textbf{150 Hz}. In practice, CIK was never the runtime bottleneck. The dominant sources of latency were serial communication with the LEAP hand and AprilTag-based object pose estimation. We provide more details of the CIK algorithm below.

\paragraph{Joint Classification.}
We classify every joint of a robotic hand into one of two categories according to its dominant geometric effect on the fingertip position in the sphere coordinate frame. Classification is performed once per hand from its URDF in a base open-hand configuration. For each joint, we sweep it across its full range of motion while keeping all other joints fixed, and we record the resulting changes in the fingertip's spherical coordinates $(\theta, \phi, r)$ via forward kinematics. In cases where a joint's rotation axis points toward the fingertip, we partially flex the remaining joints of the finger and repeat the sweep to determine its dominant effect. Joints are then categorized as follows:

\begin{itemize}
    \item \textbf{Lateral joints} are those that predominantly affect the azimuthal angle $\theta$ of the fingertip, producing side-to-side finger motion.
    \item \textbf{Encompassing joints} are those that most strongly influence the radial distance $r$ and polar angle $\phi$ of the fingertip, allowing the finger to conform to the sphere surface.
\end{itemize}

\begin{figure}[h]
\centering
\includegraphics[width=\linewidth]{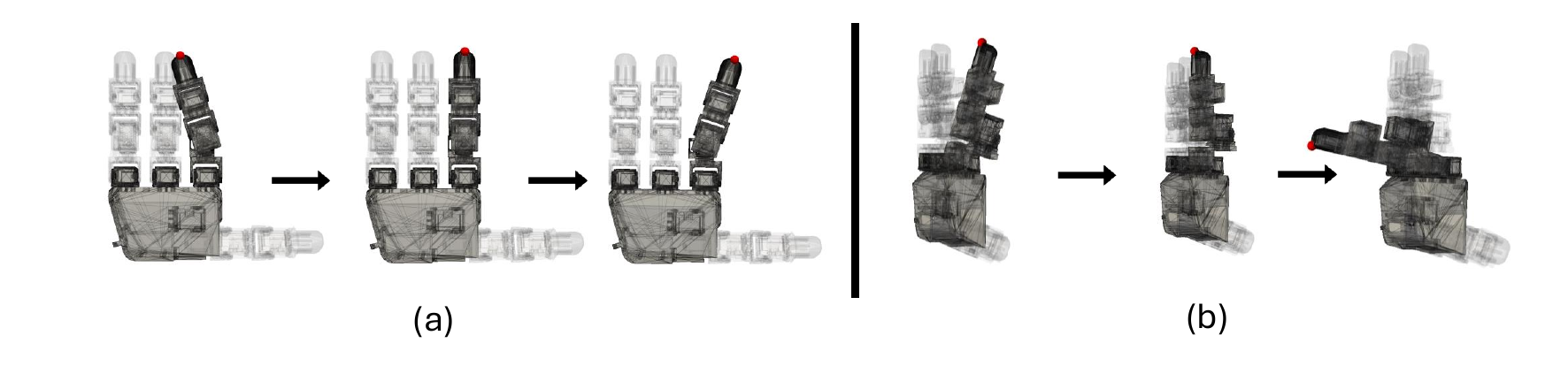}
\vspace{-2mm}
\caption{Illustration of the joint classification procedure in the Cascade Inverse Kinematics (CIK) algorithm. (a) Classification of lateral joint based on their effect on the fingertip azimuthal angle $\theta$. (b) Classification of encompassing joints based on their effect on the fingertip radial distance $r$ and polar angle $\phi$.}
\label{fig:joint_class}
\vspace{-2mm}
\end{figure}

This joint classification procedure is illustrated in Fig.~\ref{fig:joint_class}, where a joint is automatically classified according to the effect of the joint change on the position of the fingertip. Because the fingers are kinematically independent in the UHAS formulation, lateral and encompassing joints are solved sequentially but \emph{independently for each finger}.

\paragraph{Lateral Joint Mapping via Lookup Table.}
To enable fast lateral finger control, we precompute a lookup table for every lateral joint with the following steps:
\begin{enumerate}
    \item Sweep the lateral joint across its full range of motion in uniform steps.
    \item For each sampled value, fix the lateral joint and solve the encompassing joints on the \emph{undeformed} reference sphere.
    \item Record the resulting azimuthal angle $\theta_{\text{fingertip}}$ of the corresponding fingertip (surface point) on the canonical sphere frame.
    \item Store the mapping: lateral joint value $q_{\text{lateral}} \mapsto \theta_{\text{fingertip}}$.
\end{enumerate}

During inference, the policy outputs a lateral deformation $\Delta\theta$ for the driving plane aligned with the finger. We compute the fingertip target angle
\[
\theta_{\text{fingertip}} = \theta_{\text{initial}} + \Delta\theta,
\]
where $\theta_{\text{initial}}$ corresponds to the neutral (middle-of-range) pose of the finger. The precomputed lookup table directly yields the target lateral joint value $q_{\text{lateral}}$. The table is built offline once per hand and provides constant-time lateral solving at runtime.

\paragraph{Encompassing Joint Cascade.}
After the lateral joints have been set, we solve the encompassing joints sequentially in proximal-to-distal order along each finger’s kinematic chain. For each encompassing joint $i$, we first use forward kinematics with the already computed parent joint values to transform the target sphere surface points associated with joint $i$ and all its descendant links into the local coordinate frame of joint $i$. We then directly compute the joint angle that places both the current joint’s surface correspondences and those of its children onto the deformed sphere surface, as illustrated in Fig.~\ref{fig:CIK}. Because each joint is solved exactly once in a single forward pass with no iterative numerical optimization, the cascade remains extremely lightweight. This cascaded procedure progressively conforms the finger geometry to the target sphere deformation while preserving kinematic feasibility. The final joint configuration $\mathbf{q}$ is the output of the complete CIK algorithm.

\paragraph{Sphere Controller.}
By combining the compact sphere-deformation action space with CIK we obtain the complete \emph{Sphere Controller}. At every policy timestep the actor predicts driving-plane rotations ($\Delta\theta$) and driving-vector displacements ($\Delta r$). These parameters define a deformed sphere that is passed to CIK, which returns embodiment-specific joint positions $\mathbf{q}$ for the low-level controller. The lightweight sequential structure of CIK enables real-time execution across diverse robotic hands while remaining fully decoupled from any particular morphology. This design is central to the strong zero-shot transfer and rapid finetuning performance reported in Section~\ref{sec:exp}.

\section{Policy Details}
\label{appendix:training}
We provide implementation details of the reinforcement learning policies trained using the Unified Hand Action Space (UHAS). We describe the homogeneous observation space and the sphere-based action parameterization that enable effective cross-embodiment learning as well as training hyperparameters, domain randomization, and reward details in Appendix~\ref{appendix:training_rl}. 

A key requirement for training a single policy across robotic hands with different kinematic structures is the normalization of both \emph{observation and action spaces}. Standard observation formulations for in-hand manipulation include goal rotation, object pose, object linear and angular velocities, the quaternion difference to the target orientation, joint positions and velocities, and previous actions. However, raw joint values cannot be used directly because their semantic meaning and numerical ranges vary significantly across embodiments, hindering cross-embodiment transfer. We therefore represent hand configuration through points defined via surface correspondences on the finger chains (Section~\ref{sec:surface_correspondence}).

\paragraph{Sphere Observation Space.}
We construct homogeneous observations from points sampled along each finger rather than from raw joint values. For every finger, we first discretize the kinematic chain into 7 equally spaced points from root to fingertip. The parent joint of each sampled point is determined by its closest surface correspondence on the canonical sphere (Section~\ref{sec:surface_correspondence}). Positions of these points under the current joint configuration are obtained via forward kinematics. Linear and angular velocities of the points are computed using the corresponding Jacobians evaluated at the current joint configuration and joint velocities.

For all policies presented in the main paper we retain only two points per finger: the middle point and the fingertip. This choice provides a compact yet informative representation of finger configuration while remaining consistent across hands with different numbers of fingers. For 4-finger hands we duplicate the ring-finger observations to preserve dimensional consistency with 5-finger embodiments. All point positions and velocities are expressed in the canonical sphere coordinate frame (Fig.~\ref{fig:sphere_construction}) and normalized by the hand-specific sphere radius $r$. An illustration of the homogeneous observation points across different hands is shown in Fig.~\ref{fig:Homogeneous observations}. A detailed ablation on the number of observation points is provided in Appendix~\ref{appendix:ablations}.

\begin{figure}[h]
\centering
\includegraphics[width=\linewidth]{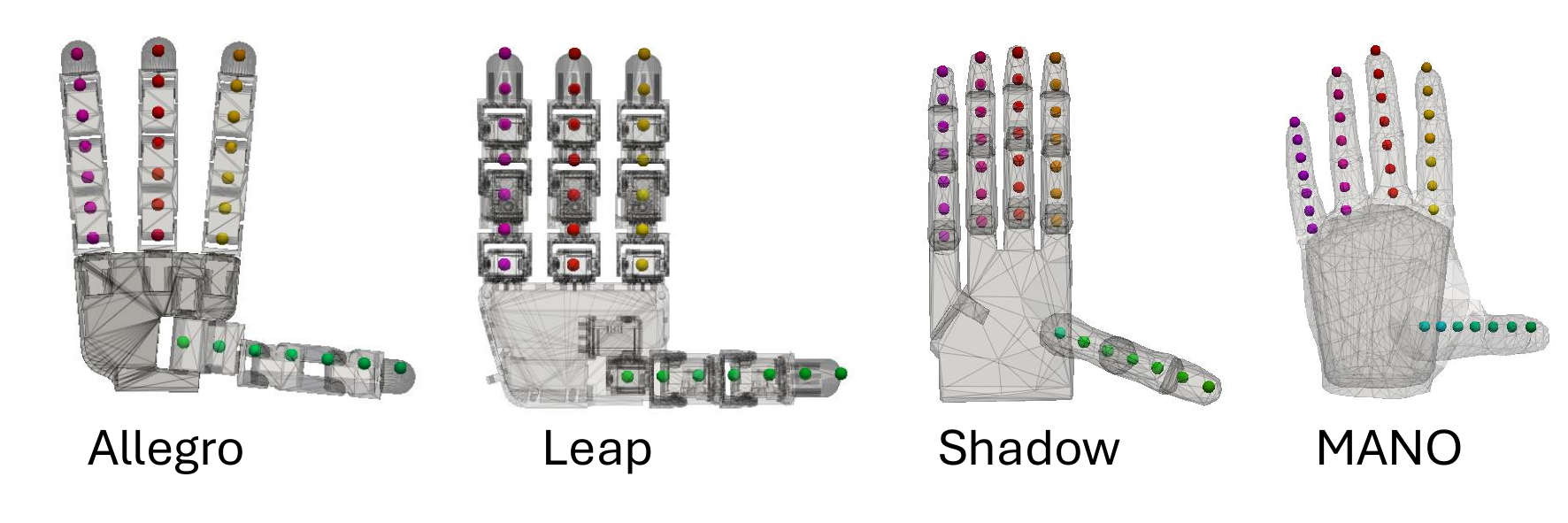}
\vspace{-2mm}
\caption{Homogeneous observations across different robotic hands.}
\label{fig:Homogeneous observations}
\vspace{-2mm}
\end{figure}

\paragraph{Sphere Action Space.}
Actions are defined in the sphere deformation space introduced in Section 3.3. Each action consists of lateral deformations $\Delta\theta$ applied to a set of driving planes and radial deformations $\Delta r$ applied to driving vectors within those planes. We attach one driving plane per fingertip and align its initial azimuthal angle with the corresponding fingertip $\theta$ on the canonical sphere. To enable a unified policy across both 4-finger and 5-finger hands while preserving independent actuation of the fifth finger, we introduce an additional driving plane at the ring-finger azimuthal position for all 4-finger embodiments. When computing deformations in the ring finger, the $\Delta\theta$ and $\Delta r$ actions of the duplicated plane are averaged before interpolation.

Lateral actions are bounded to the extremal values stored in the precomputed lookup tables of each hand. We employ two driving vectors per plane, positioned at equally spaced polar angles $\phi = 60^\circ$ and $\phi = 120^\circ$, with radial displacements defined over the interval $[-2, 2]$. Prior to invoking the Cascade Inverse Kinematics (CIK) algorithm, we discard all sphere points whose radial coordinate $r$ is negative. If an encompassing joint has no reachable points remaining after this filtering step, the joint is commanded to its fully closed configuration. This design choice was made after observing that permitting negative radial deformations substantially improves performance on the highly dynamic cube reorientation task, as it allows the policy to close fingers rapidly during aggressive reposing maneuvers. An ablation study on the number of driving vectors per plane is provided in Appendix~\ref{appendix:ablations}.

\subsection{Training}
\label{appendix:training_rl}

\paragraph{Hyperparameters.}
All models were trained on a single NVIDIA A5000 GPU using the RSL-RL implementation of Proximal Policy Optimization (PPO) \cite{schwarke2025rslrl}. Training is performed in the custom NVIDIA Isaac Lab Cube Reposing environment \cite{mittal2025isaaclab} running on Isaac Sim 4.5.0 with the PhysX physics simulator. The complete set of PPO training hyperparameters is summarized in Table~\ref{tab:ppo_hyperparameters}.

\begin{table}[h]
\centering
\caption{PPO training hyperparameters.}
\label{tab:ppo_hyperparameters}
\vspace{-2mm}
\begin{tabular}{ll}
\toprule
\textbf{Hyperparameter}              & \textbf{Value}                  \\
\midrule
\multicolumn{2}{l}{\textit{Runner Configuration}} \\
\midrule
Num. steps per environment           & 16                              \\
Empirical normalization              & True                            \\
\midrule
\multicolumn{2}{l}{\textit{Policy Network (Actor \& Critic)}} \\
\midrule
Hidden layer dimensions              & $[512, 512, 256, 128]$          \\
Activation function                  & ELU                             \\
Initial action noise std.            & $1.0$                           \\
\midrule
\multicolumn{2}{l}{\textit{PPO Algorithm}} \\
\midrule
Learning rate                        & $5.0 \times 10^{-4}$            \\
Learning rate schedule               & adaptive                        \\
Clip parameter                       & $0.2$                           \\
Entropy coefficient                  & $0.005$                         \\
Num. learning epochs                 & $5$                             \\
Num. mini-batches                    & $4$                             \\
Value loss coefficient               & $1.0$                           \\
Use clipped value loss               & True                            \\
Discount factor ($\gamma$)           & $0.99$                          \\
GAE parameter ($\lambda$)            & $0.95$                          \\
Desired KL divergence                & $0.016$                         \\
Max. gradient norm                   & $1.0$                           \\
\bottomrule
\end{tabular}
\vspace{-2mm}
\end{table}

\paragraph{Reward Function.}
We adopt the reward formulation from the original NVIDIA Isaac Lab Reposing Cube environment. To discourage embodiment-specific exploitation—such as policies that rely predominantly on lateral joint motion while underutilizing encompassing joints—we augment the original reward with two additional regularization terms. These terms penalize deviations of the lateral and encompassing joint positions from a reference joint configuration. Let $d$ denote the object-to-goal distance and $r_{\text{rot}}$ the orientation alignment reward. Let $p_{\text{lat}}$ and $p_{\text{rad}}$ denote the lateral and encompassing joint position penalties, respectively. The scalar reward at each timestep is then computed as
\begin{equation}
    r = w_d \, d + w_r \, r_{\text{rot}} + w_{\text{lat}} \, p_{\text{lat}} + w_{\text{rad}} \, p_{\text{rad}} + b_{\text{success}} + p_{\text{fall}},
\end{equation}
where $b_{\text{success}}$ is a large positive bonus awarded upon reaching the goal within tolerance of 0.1 radians, and $p_{\text{fall}}$ is a negative penalty applied when the object falls. The reward scales used during training are reported in Table~\ref{tab:reward_scales}.

\begin{table}[h]
\centering
\caption{Reward scales used during training.}
\label{tab:reward_scales}
\vspace{-2mm}
\begin{tabular}{ll}
\toprule
\textbf{Reward Term}                              & \textbf{Scale}          \\
\midrule
Object-to-goal distance ($w_d$)                   & $-10.0$                 \\
Orientation alignment ($w_r$)                     & $1.0$                   \\
Reach goal bonus ($b_{\text{success}}$)           & $250$                   \\
Fall penalty ($p_{\text{fall}}$)                  & $-100$                  \\
Lateral joint position penalty ($w_{\text{lat}}$) & $-0.016$                \\
Encompassing joint position penalty ($w_{\text{rad}}$) & $-0.004$           \\
\bottomrule
\end{tabular}
\vspace{-2mm}
\end{table}
\paragraph{Domain Randomization.}

\begin{table}[h]
\centering
\caption{Domain randomization ranges applied during training.}
\label{tab:domain_randomization}
\vspace{-2mm}
\small
\begin{tabular}{ll}
\toprule
\textbf{Parameter}                              & \textbf{Randomization Range}          \\
\midrule
\multicolumn{2}{l}{\textit{Object Properties}} \\
\midrule
Object scale                                    & $[0.9, 1.1] \times$ nominal         \\
Object mass                                     & $[0.8, 1.2] \times$ nominal         \\
Object static friction                          & $[0.2, 0.3]$                        \\
Object dynamic friction                         & $[0.15, 0.25]$                      \\
\midrule
\multicolumn{2}{l}{\textit{Robot Properties}} \\
\midrule
Robot mass                                      & $[0.9, 1.1] \times$ nominal         \\
Robot static / dynamic friction                 & $[0.75, 1.0]$                       \\
Joint friction                                  & $[0.9, 1.1] \times$ nominal         \\
Joint armature                                  & $[1.00, 1.05] \times$ nominal       \\
Joint effort limits                             & $[0.9, 1.1] \times$ nominal         \\
Joint stiffness                                 & $[0.75, 1.25] \times$ nominal       \\
Joint damping                                   & $[0.75, 1.25] \times$ nominal       \\
\midrule
\multicolumn{2}{l}{\textit{Hand Pose \& Action Space}} \\
\midrule
Hand base inclination                           & $15^\circ \pm 5^\circ$              \\
Driving vector azimuthal angle ($\phi$)         & $\pm 15^\circ$                      \\
\bottomrule
\end{tabular}
\vspace{-2mm}
\end{table}

We utilize domain randomization during policy training. These randomization strategies proved beneficial for improving cross-embodiment generalization. Randomizing the hand base inclination discouraged policies from relying on specific cube-palm interaction dynamics, such as assuming the cube would slide or settle into particular regions of the palm. Object scale randomization helped mitigate finger entrapment issues arising from differences in finger geometry across embodiments. Additionally, randomizing the driving vector azimuthal angle $\theta$, joint effort limits, and PD gains altered the in-distribution dynamics of each hand which improved generalization, particularly across certain hands. An ablation study examining the contribution of these randomization components is provided in Appendix~\ref{appendix:ablations}. Table~\ref{tab:domain_randomization} shows the details of the domain randomization used for training. Note that the domain randomization strategy used for real-world deployment differed from the one described in this section. Details of real-world deployment are provided in Appendix~\ref{appendix:sys_ID}.

\section{Cube Pose Estimation}
\label{appendix:cube_pose}

We track a 3D-printed cube based on an open source design~\cite{park2026aprilcube}. Each of its six faces carries four distinct \texttt{tag36h11} AprilTags~\cite{apriltag} (24 tags total). One or two Intel RealSense cameras each stream a rectified infrared image at $848\times480$ pixels and 60 Hz. A second camera viewpoint increases the number of visible tags on the cube that are not occluded by the hand. We also placed the hand in a well-lit location allowing for shorter exposure time to reduce motion blur during fast reorientation movements.

Each visible tag is localized independently via perspective-n-point pose estimation using its four corners and the camera intrinsics. Detected tags are only published if the Hamming decoded value is zero and the decision margin is at least 50. A separate standalone tag is placed at a known location on the base mount and serves as a common reference frame.

In the dual-camera configuration, messages timestamped within 100 ms of each other are paired. The transformation from the secondary to the primary camera frame is estimated from tags visible in both camera views. In the case of duplicate tags which both cameras see, the tag with the higher detection margin will be kept. Given the final set of visible tags, the cube pose is calculated by chaining the camera-to-tag pose with the calibrated cube-to-tag transform. Estimates whose translation deviates by more than 10 mm from the median or whose orientation deviates by more than 0.075 rad from a consensus quaternion are dropped. The remaining estimates are fused by their mean position and hemisphere-aligned quaternion averaging. The cube pose is published only when at least three separate tags are visible.

\section{System Identification}
\label{appendix:sys_ID}

To enable reliable sim-to-real transfer, we performed system identification on the complete real-world hardware setup, including the LEAP Hand, its actuators, and the communication interface. Due to the limited serial communication bandwidth of the LEAP Hand, we operate the policy at a control frequency of 20 Hz during training and real-world deployment.

% We model the servo control law as current $I(t) = K_p \Delta\theta(t) - K_d \Delta\dot{\theta}(t)$, where $\Delta\theta(t)$ is the joint error, and $K_p$ and $K_d$ are the PD control gains. However, directly mapping this formulation to the implicit PD torque actuators in simulation did not produce motion profiles consistent with real-world behavior. To address this discrepancy, we identified the effective proportional and derivative gains $K_p$ and $K_d$ directly from the physical LEAP Hand. We collected data by commanding random target positions under varying gain settings and solved for the gains that best reproduced the observed joint position, velocity and current readings. The identified values differed substantially from the manufacturer specifications, particularly in the damping term. We attribute this difference to the current-based position control mode used by the motors. \yu{Do you use these identified values in simulation training? We need to write about what the result of the system identification.}

We model the servo control law of the LEAP Hand as $\text{current}(t) = K_p \Delta\theta(t) - K_d \Delta\dot{\theta}(t)$, where $\Delta\theta(t)$ denotes the joint position error. Directly mapping this current-based formulation to the implicit PD torque actuators available in simulation did not reproduce the motion observed on the physical robot. To resolve this mismatch, we performed system identification on the LEAP Hand by commanding random target positions while varying the commanded $K_p$ and $K_d$ gains. From the recorded joint position, velocity, and current trajectories, we solved for the effective gains that best matched the recorded data. The identified values were subsequently used when training policies intended for real-world deployment.

The system identification revealed that the LEAP Hand motors behave as a nearly undamped system, with damping having a negligible effect on the observed dynamics. Furthermore, the effective proportional gain $K_p$ differed from the manufacturer specifications: approximately $0.0786~\text{Nm/rad}$ per 100 motor units. The effective derivative gain $K_d$ was found to be approximately $0.0014~\text{Nm/(rad/s)}$ per 100 motor units. We attribute these discrepancies primarily to the current-based position control mode employed by the motors.

By monitoring the motor state during operation, we observed that the LEAP Hand reliably enforces its internal velocity limits. Accordingly, we introduced velocity limits in simulation when training models intended for real-world deployment. These limits were themselves randomized during training to improve robustness.

\subsection{Training for Real-World Deployment}

For models deployed on the physical robot, we increased the range of domain randomization compared to simulation-only training. In addition to the randomization parameters described in Table~\ref{tab:domain_randomization}, we randomized the velocity limits applied in simulation. We also adopted an asymmetric actor-critic architecture: the actor receives only positional information of the object and hand joints, while the critic has access to the full state, including linear and angular velocities of the object and hand. This design encourages the actor to learn policies that rely primarily on position feedback, which is more reliable under real-world sensing and communication noise.

\subsection{Deployment on the Physical LEAP Hand}

During real-world operation, we observed that serial communication with the LEAP Hand was unreliable, with frequent missed reads and writes. To mitigate this, we implemented additional software-level communication management to improve reliability. Despite these measures, certain motor state readings consistently failed through the manufacturer API.

We further limited the range of several joints on the physical hand. This was necessary because the joints exhibited overshoot and because the structural components of the LEAP Hand are relatively deformable under load. We also imposed a lower velocity limit than the hardware maximum to reduce overshooting observed during fast motions.

Finally, we observed that the fingers of the LEAP Hand gradually became loose after repeated use. To improve mechanical stability, we designed and 3D-printed a reinforced base that secures the root joints of the index, middle, and ring fingers using additional screws. We also observed significant deformation at the thumb root joint and therefore designed a custom attachment for it. Both the reinforced finger base and the thumb attachment will be released publicly alongside the paper.

\section{Additional Ablation Studies}
\label{appendix:ablations}

We present additional ablation studies on two key design choices in our method: the number of homogeneous observation points per finger and the number of driving planes in the Unified Hand Action Space. In addition, we present an ablation study on domain randomization.

\paragraph{Ablation on Observation Points.}
We ablate the number of homogeneous observation points per finger. Fig.~\ref{fig:Homogeneous observations} illustrates the full set of candidate points distributed along each finger. From these points, we select different subsets as observations: one point at the fingertip, two points consisting of the finger midpoint and fingertip, or three and four equally spaced points along the finger. All models were trained amd tested on the four hands using one NVIDIA A5000 GPU while keeping all other components fixed. As reported in Table~\ref{tab:observation_ablation}, increasing the number of observation points provides only marginal improvements in success rate and average consecutive reorientations. These gains become negligible when domain randomization is applied, while training time and inference cost increase. We therefore use two observation points per finger (midpoint and fingertip) in our final models, as this configuration achieves a favorable trade-off between performance and computational efficiency. Additional details regarding the construction of the homogeneous observations are provided in Appendix~\ref{appendix:training}.

\begin{table}[h]
\centering
\caption{Ablation on the number of homogeneous observation points per finger.}
\label{tab:observation_ablation}
\vspace{-2mm}
\setlength{\tabcolsep}{6pt}
\begin{tabular}{lcccc}
\toprule
\textbf{\# Observation Points} 
& \textbf{1} 
& \textbf{2} 
& \textbf{3} 
& \textbf{4} \\
\midrule
\textbf{Success Rate} 
& 98.8
& 98.7
& 99.0
& \textbf{99.1} \\

\textbf{\# Reorientations} 
& 9.3 $\pm$ 2.1
& 9.3 $\pm$ 2.0
& 9.4 $\pm$ 1.8 
& \textbf{9.5 $\pm$ 1.8} \\

\textbf{Training Time (h)} 
& 4.9
& \textbf{4.5}
& 4.8
& 4.7 \\
\bottomrule
\end{tabular}
\vspace{-2mm}
\end{table}
\paragraph{Ablation on Driving Planes.}
Additionally, we ablate the number of driving planes in the Unified Hand Action Space (UHAS). For the non-zero-shot models, we train and evaluate each hand using either the standard 5-plane UHAS or a 4-plane variant. The 4-plane version is constructed identically to our merging procedure for 4-finger hands: we compute the average deformation of the ring and pinky planes and then interpolate the canonical sphere points as usual. For the zero-shot models, we train on the other three hands and deploy using the 4-plane UHAS (again merging the ring and pinky planes). Consequently, the 4-plane zero-shot policies move the ring and pinky fingers in a coupled manner.

As shown in Table~\ref{tab:driving_planes_ablation}, using 4 or 5 driving planes yields very similar performance. Interestingly, the Shadow hand even achieves slightly better results with 4 planes than with 5. We attribute this to the fact that the pinky finger is rarely used to solve the cube reposing task; therefore, removing its dedicated driving plane has a negligible impact on task performance.

We attempted to train models using only 3 driving planes. However, these models were unable to reliably solve the reposing task, and training failed to converge to meaningful policies. This indicates that a minimum level of action-space expressiveness is required for stable learning on this task.

It is important to note that the small difference observed between 4 and 5 planes may be specific to this reposing task, the particular reward formulation and the reference cube position used. These factors had a significant influence on hand behavior during training. Therefore, the negligible impact of removing the fifth plane should not be assumed to hold for other manipulation tasks or reward designs.

\begin{table}[h]
\centering
\caption{Ablation on the number of driving planes.}
\label{tab:driving_planes_ablation}
\vspace{-2mm}
% \scriptsize
\setlength{\tabcolsep}{8pt}
\begin{tabular}{lcc}
\toprule
\textbf{\# Driving Planes} & \textbf{4} & \textbf{5} \\
\midrule
Shadow                  & \textbf{99.5 / 9.7 $\pm$ 1.5} & 99.3 / 9.6 $\pm$ 1.6 \\
Shadow (Zero-shot)      & \textbf{86.9 / 4.8 $\pm$ 3.8} & 85.7 / 4.4 $\pm$ 3.7 \\
MANO                    & 99.6 / 9.8 $\pm$ 1.3 & \textbf{99.8 / 9.9 $\pm$ 1.0} \\
MANO (Zero-shot)        & 98.0 / 8.7 $\pm$ 2.9 & \textbf{98.1 / 8.9 $\pm$ 2.6} \\
\bottomrule
\end{tabular}
\vspace{-2mm}
\end{table}

\paragraph{Ablation on Domain Randomization.}
We evaluate the impact of domain randomization on zero-shot cross-embodiment generalization. For each hand, we train models on the other three hands and evaluate zero-shot transfer performance. We compare models trained with and without randomization of PD gains, joint effort limits, object scale, and driving vector polar angle $\phi$. Table~\ref{tab:randomization_ablation} reports the results across all four hands. Randomization during training consistently improves zero-shot transfer, particularly for hands with larger morphological differences from the training set.

\begin{table}[h]
\centering
\caption{Zero-shot performance with and without domain randomization (PD gains, effort, object scale, and vector $\phi$). Results reported as Success Rate / Average Consecutive Reorientations $\pm$ std.}
\label{tab:randomization_ablation}
\vspace{-2mm}
% \scriptsize
\setlength{\tabcolsep}{4pt}
\begin{tabular}{lcc}
\toprule
\textbf{Test Hand} & \textbf{Without Randomization} & \textbf{With Randomization} \\
\midrule
Allegro & \textbf{96.7 / 8.2 $\pm$ 3.1} & 95.3 / 7.7 $\pm$ 3.4 \\
LEAP    & 95.3 / 7.6 $\pm$ 3.4 & \textbf{95.5 / 7.7 $\pm$ 3.5} \\
Shadow  & 80.1 / 3.6 $\pm$ 3.6 & \textbf{85.7 / 4.4 $\pm$ 3.7} \\
MANO    & 97.1 / 8.4 $\pm$ 2.9 & \textbf{98.1 / 8.9 $\pm$ 2.6} \\
\bottomrule
\end{tabular}
\vspace{-2mm}
\end{table}

\section{Additional Experimental Results}
\label{appendix:results}

We present additional experimental results that complement the main paper. We first analyze one-to-many zero-shot generalization by training policies on a single source hand and evaluating them on all other target hands. We then report real-world results on the Allegro Hand \cite{allegrohand}.

\subsection{One-to-Many Generalization in Simulation}

We train a policy on a single source hand and evaluate its zero-shot performance on all other target hands. This setting reveals how well a policy learned on one embodiment can transfer without any finetuning or exposure to other hands during training. Table~\ref{tab:one_to_many} summarizes the results.

\begin{table}[h!]
\centering
\caption{One-to-Many zero-shot generalization. Rows indicate the source hand used for training; columns indicate the target hand used for evaluation.}
\label{tab:one_to_many}
\vspace{-2mm}
\setlength{\tabcolsep}{4pt}
\begin{tabular}{lcccc}
\toprule
\textbf{Source \textbackslash{} Target} & \textbf{Allegro} & \textbf{LEAP} & \textbf{Shadow} & \textbf{MANO} \\
\midrule
Allegro & \textbf{99.1 / 9.6 $\pm$ 1.7} & 55.4 / 1.1 $\pm$ 1.5 & 8.7 / 0.1 $\pm$ 0.3 & 17.5 / 0.0 $\pm$ 0.13 \\
LEAP    & 95.3 / 7.8 $\pm$ 3.3 & \textbf{99.7 / 9.8 $\pm$ 1.1} & 65.8 / 1.8 $\pm$ 1.9 & 87.0 / 4.7 $\pm$ 3.7 \\
Shadow  & 35.6 / 0.4 $\pm$ 0.7 & 59.4 / 1.6 $\pm$ 2.2 & \textbf{99.3 / 9.6 $\pm$ 1.6} & 97.6 / 8.7 $\pm$ 2.7 \\
MANO    & 33.0 / 0.4 $\pm$ 0.68 & 31.0 / 0.4 $\pm$ 0.67 & 36.2 / 0.5 $\pm$ 0.9 & \textbf{99.8 / 9.9 $\pm$ 1.0} \\
\bottomrule
\end{tabular}
\vspace{-2mm}
\end{table}

The results highlight two important phenomena. First, policies trained on a single hand frequently develop \emph{exploitative behaviors} that are highly specific to that embodiment. For example, the Allegro-trained policy relies heavily on the hand’s distinctive lateral joints to perform cube rotations. Because Shadow and MANO lack equivalent lateral actuation, this policy fails almost completely on those hands. Interestingly, it retains partial transfer to the LEAP Hand, likely because the LEAP Hand’s lateral joints can produce somewhat similar motion when the fingers are flexed.

Second, hands with larger joint ranges of motion and workspace tend to generalize better to other embodiments. The LEAP Hand, which possesses the largest range of motion among the four, achieves the strongest overall zero-shot performance when used as the source. Similarly, the Shadow Hand generalizes reasonably well to MANO, but the reverse direction (MANO → Shadow) performs significantly worse, despite both hands having five fingers. This asymmetry suggests that greater kinematic flexibility during training enables the policy to learn more transferable behaviors, whereas policies trained on more constrained hands tend to overfit to their specific joint limits and motion patterns.

These findings raise an interesting question for future work: whether explicitly limiting the range of motion or action space during training on high-degree-of-freedom hands could improve robustness, or whether the current asymmetry is inherent to cross-embodiment transfer.

\subsection{Real-World Results on the Allegro Hand}

\begin{figure}[h]
\centering
\includegraphics[width=0.5\linewidth]{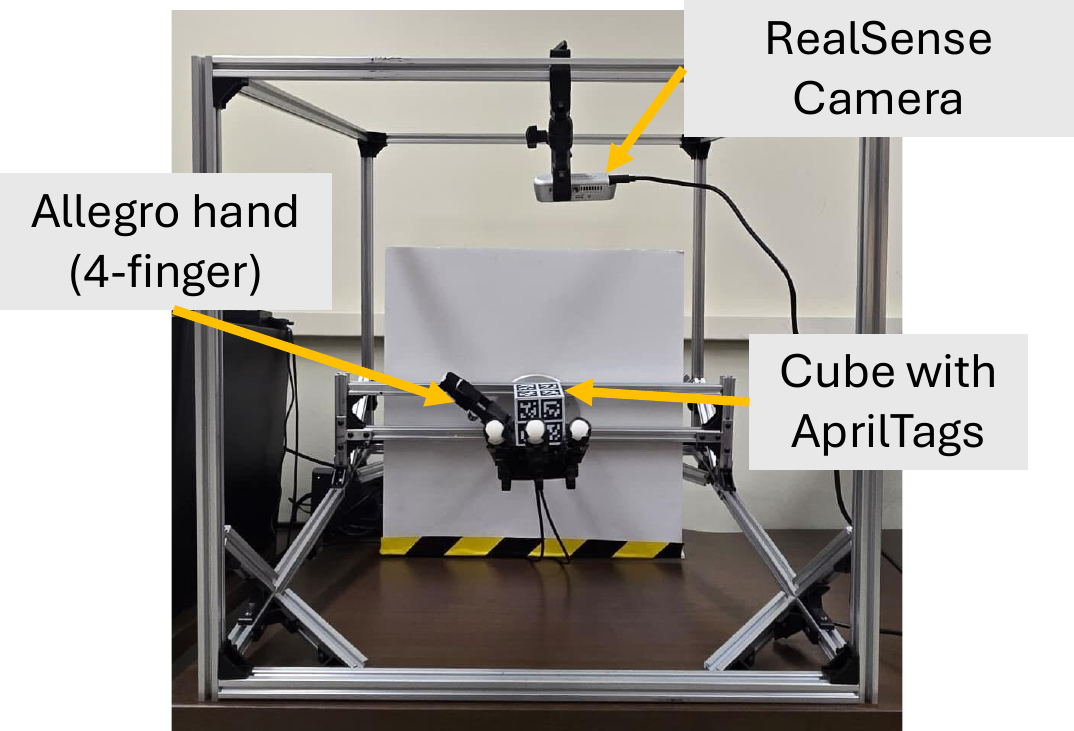}
\caption{Illustration of our real-world setup for the Allegro hand}
\label{fig:allegro_real_world}
\vspace{-2mm}
\end{figure}

We deploy our method on a physical Allegro Hand~\cite{allegrohand} along with a cube pose estimator based on AprilTags. The real-world setup is shown in Fig.~\ref{fig:allegro_real_world}. We evaluate in-hand cube reorientation over 10 independent trials per method. In each trial, the policy runs continuously until the cube falls off the hand. Table~\ref{tab:allegro_real_world} reports the number of consecutive successful reorientations achieved before failure across trials, along with the mean. 

\begin{table}[h]
\centering
\caption{Real-world in-hand cube reorientation on the Allegro Hand over 10 independent trials. Entries report the number of consecutive successful reorientations before failure.}
\label{tab:allegro_real_world}
\vspace{-2mm}
\scalebox{0.8}{
\begin{tabular}{lcccccccccc c}
\toprule
\textbf{Method} & \textbf{1} & \textbf{2} & \textbf{3} & \textbf{4} & \textbf{5} & \textbf{6} & \textbf{7} & \textbf{8} & \textbf{9} & \textbf{10} & \textbf{MEAN} \\
\midrule
% Baseline (Joint Control) & XX & XX & XX & XX & XX & XX & XX & XX & XX & XX & X.X \\ 
% Abhijit: Commented the baseline as I don't think we have time to finish that
UHAS (Zero-Shot)         & 0 & 1 & 1 & 1 & 0 & 1 & 0 & 0 & 2 & 2 & 0.8 \\
UHAS (Trained on Multi-Hand)         & 3 & 0 & 8 & 1 & 2 & 1 & 1 & 2 & 1 & 2 & \textbf{2.1} \\
UHAS (Trained on Allegro Hand) & 4 & 0 & 2 & 2 & 2 & 3 & 4 & 1 & 0 & 3 & \textbf{2.1} \\
\bottomrule
\end{tabular}
}
%\vspace{-2mm}
\end{table}

The Allegro-trained model achieves the highest real-world performance with a mean of 2.1 consecutive reorientations, tied with the multi-hand trained model. The zero-shot policy, however, performs substantially worse (mean 0.8), revealing a notable sim-to-real gap for cross-embodiment transfer, consistent with our LEAP Hand results. We also observe high variance across trials, with some runs reaching 8 reorientations while others fail immediately.

\end{document}